\documentclass[letterpaper, 10pt, conference]{ieeeconf}  

\IEEEoverridecommandlockouts   
\usepackage[utf8]{inputenc} 
\usepackage[T1]{fontenc}    

\usepackage{url}            
\usepackage{booktabs}       
\usepackage{amsfonts}       
\usepackage{amsmath,amssymb}
\usepackage{nicefrac}       
\usepackage{microtype}      
\usepackage[dvipsnames]{xcolor} 
\usepackage{graphicx}
\usepackage{caption}
\usepackage{subcaption}
\usepackage{wrapfig}
\usepackage{placeins}
\usepackage{tabularx}
\usepackage{multirow}
\usepackage{gensymb}
\usepackage{siunitx}
\usepackage{pifont}         
\usepackage{xspace}
\usepackage{algorithm}
\usepackage[algo2e]{algorithm2e}
\usepackage{algpseudocode}

\usepackage{hyperref}
\hypersetup{
    colorlinks=true,
    linkcolor=blue,
    citecolor=blue,
    urlcolor=blue,
    pdfborder={0 0 0},
}

\newcommand{\cmark}{\ding{51}} 
\newcommand{\xmark}{\ding{55}} 

\def\figref#1{Fig.~\ref{#1}}

\def\tabref#1{Tab.~\ref{#1}}

\def\eqref#1{Eq.~\ref{#1}}

\definecolor{nvgreen}{RGB}{65,125,0}
\newcommand{\yescell}{\textcolor{nvgreen}{\cmark}}
\newcommand{\nocell}{\textcolor{red}{\xmark}}

\makeatletter
\providecommand{\bstctlcite}{\@ifnextchar[{\@bstctlcite}{\@bstctlcite[@auxout]}}
\def\@bstctlcite[#1]#2{\@bsphack
  \@for\@citeb:=#2\do{%
    \edef\@citeb{\expandafter\@firstofone\@citeb}%
    \if@filesw\immediate\write\csname #1\endcsname{\string\citation{\@citeb}}\fi}%
  \@esphack}

\let\standardthebibliography\thebibliography
\renewcommand{\thebibliography}[1]{\standardthebibliography{#1}\scriptsize}
\makeatother

\newcommand{\figtop}{{\em (Top)}\xspace}
\newcommand{\figbottom}{{\em (Bottom)}\xspace}

\newcommand{\benchmark}{UMI-Loc-Bench-v1\xspace}

\newcommand{\ourmethod}{\textcolor{nvgreen}{OpenJaws}\xspace}

\newif\ifsubmission
\submissionfalse

\newif\iffullrefs
\fullrefstrue

\newcommand{\camerareadyauthors}{}

\title{\LARGE \bf
\raisebox{-0.3\height}{\includegraphics[height=1.6em]{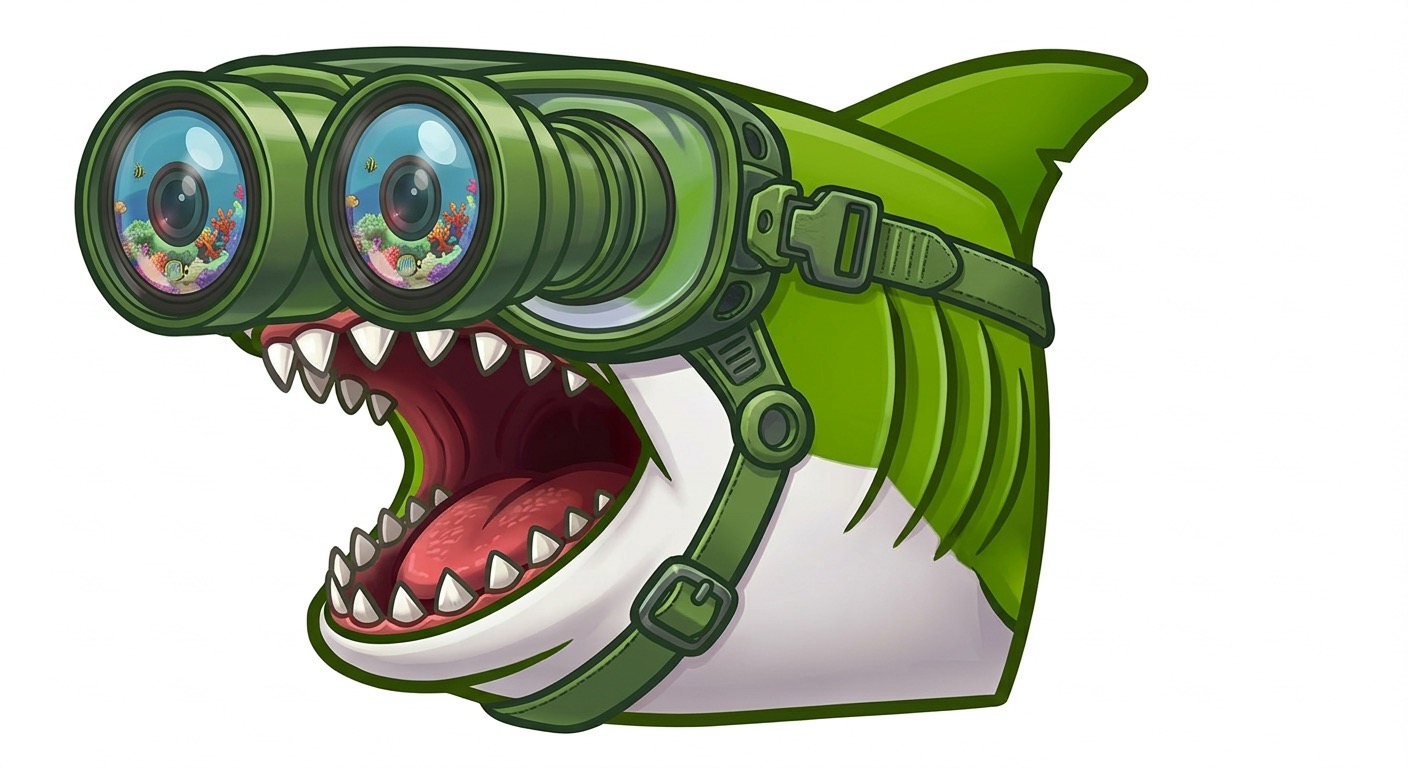}}\hspace{-0.35em}%
\textcolor{nvgreen}{\textbf{\ourmethod:}} An Open Hardware UMI Framework \\ with Stereo Localization for Generalist Manipulation}

\ifdefined\camerareadyauthors
  \author{Adithyavairavan Murali, Yash Oza, Karanbir Chahal, Neel Jawale, Silas De Vries,\\
          Hesam Rabeti, Erwin Coumans, Balakumar Sundaralingam, Stan Birchfield\\[0.1em]
          \raisebox{-0.15in}[0pt][0pt]{\includegraphics[height=0.28in,trim=120bp 20bp 120bp 20bp,clip]{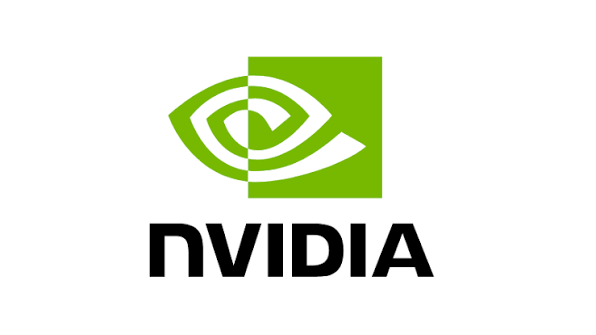}}\\[0.18in]
          \href{https://open-jaws.github.io/}{\textcolor{nvgreen}{\textbf{https://open-jaws.github.io/}}}}
\else
  \ifsubmission
    \author{Anonymous Author(s)}
  \else
    \author{Author One$^{1}$, Author Two$^{1}$, and Author Three$^{1}$%
    \thanks{$^{1}$Affiliation, Address, Country
            {\tt\small email@example.com}}%
    }
  \fi
\fi

\IEEEaftertitletext{%
  \ifdefined\camerareadyauthors\vspace{-0.12in}\fi
  \begin{minipage}{\textwidth}
    \centering
    \includegraphics[width=\textwidth]{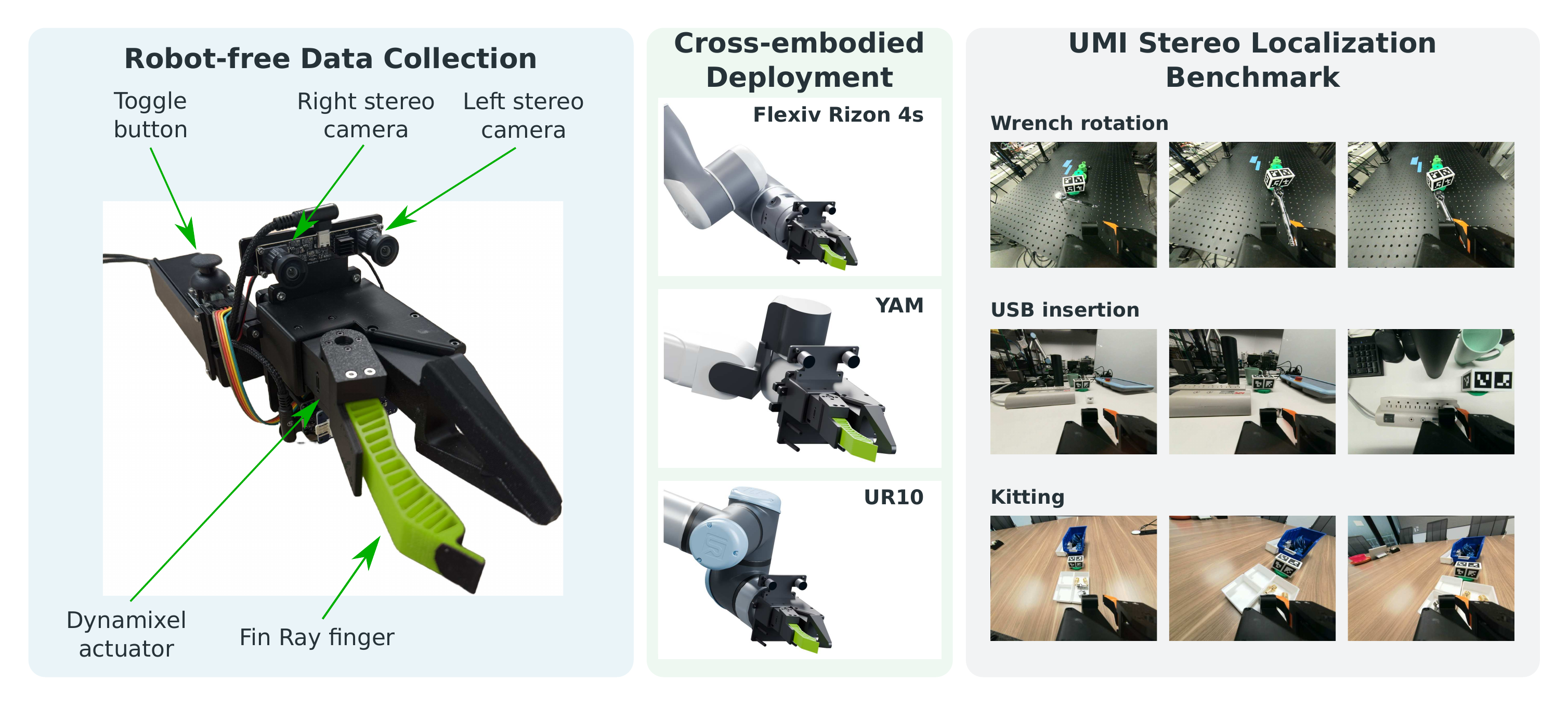}
    \captionof{figure}{\textbf{\ourmethod system overview.} Robot-free demonstrations are collected with an actuated, stereo-equipped gripper (\emph{left}), zero-shot transferred across robot embodiments through a standardized mounting interface (\emph{center}), and evaluated on \benchmark using diverse manipulation sequences (\emph{right}).}
    \label{fig:teaser}
  \end{minipage}%
  \vspace{0.5\baselineskip}%
}

\begin{document}

\bstctlcite{BSTcontrol}

\maketitle
\thispagestyle{empty}
\pagestyle{empty}   

\begin{abstract}
Handheld data-collection interfaces such as the Universal Manipulation Interface (UMI) allow manipulation demonstrations to be gathered in the wild without a robot in the loop, but the quality of the resulting action labels is bounded by the accuracy of vision-based localization. Existing systems recover end-effector pose from monocular visual-inertial odometry, which is subject to scale ambiguity and drift, or delegate tracking to closed-source modules and external infrastructure that reintroduce cost and setup burden. We present \ourmethod, a fully open hardware and software framework for sensorized UMI that improves localization and simplifies deployment across robot platforms. \ourmethod pairs a modular, 3D-printed gripper exposing a standardized ISO adapter with a low-cost stereo camera to recover metric camera trajectories. We further introduce \benchmark, a benchmark quantifying real-world localization accuracy for handheld data-collection devices, and show that stereo localization substantially outperforms monocular alternatives. Finally, we demonstrate that policies trained on data collected with \ourmethod can be deployed on a real robot across multiple tasks. Hardware, software, and data are released under a commercially permissive license. Code, data, and videos can be found at the website: \mbox{\href{https://open-jaws.github.io/}{\textcolor{nvgreen}{https://open-jaws.github.io/}}}
\end{abstract}

\section{Introduction}
\label{sec:introduction}

We have witnessed a cambrian explosion in robot foundation models in recent years ~\cite{brohan2023rt1, brohan2023rt2, oneill2024openx, black2024pi0, nvidia2025gr00tn1, team2025geminirobotics}. These models rely on large training datasets, causing leader-follower teleoperation systems, such as ALOHA~\cite{zhao2023aloha} and GELLO~\cite{wu2024gello} to become workhorses of the industry. However, teleoperation is difficult to scale because it requires a kinematically identical copy of the robot, severely limiting in-the-wild data collection. Furthermore, predicting policy actions in joint space does not, by construction, easily generalize across different robot embodiments. 

To address this scaling bottleneck, the breakthrough work of the Universal Manipulation Interface (UMI)~\cite{chi2024umi} proposed a portable, handheld gripper for collecting data independent of the robot arm, where robot action labels are extracted offline using visual-inertial localization. A crucial (and typical) design choice is to define the policy action space in relative end-effector task space, which enables robot-arm-agnostic deployment. 

Despite UMI's significant potential, however, a large \textit{UMI-to-real} gap exists, manifesting as several engineering pain points when deploying a UMI policy. First, the quality of the policy action labels is affected by Simultaenous Localization and Mapping (SLAM), such that localization errors introduce noise into the labels, affecting policy training. Second, the original UMI required a separate, actuated parallel-jaw electric gripper at deployment time. A mount must be designed for the finger holder, and great care must be taken to ensure that the kinematics (e.g. stroke length) of the handheld device match those of the deployment gripper. One must also ensure that the relative coordinate transformation between the wrist camera and the Tool Center Point (TCP) remains consistent during both data collection and deployment.  Given this UMI-to-real gap and the associated system integration challenges, the most proficient UMI-based manipulation stacks are private and closed-source~\cite{generalistai2026,sundayai2026}.  Other recent and concurrent works~\cite{hajjahmad2026koala,seo2024legato,pollen2025grabette} co-design the handheld UMI data collection device with the deployment gripper.

\begin{figure}[t]
  \centering
  \includegraphics[width=\linewidth]{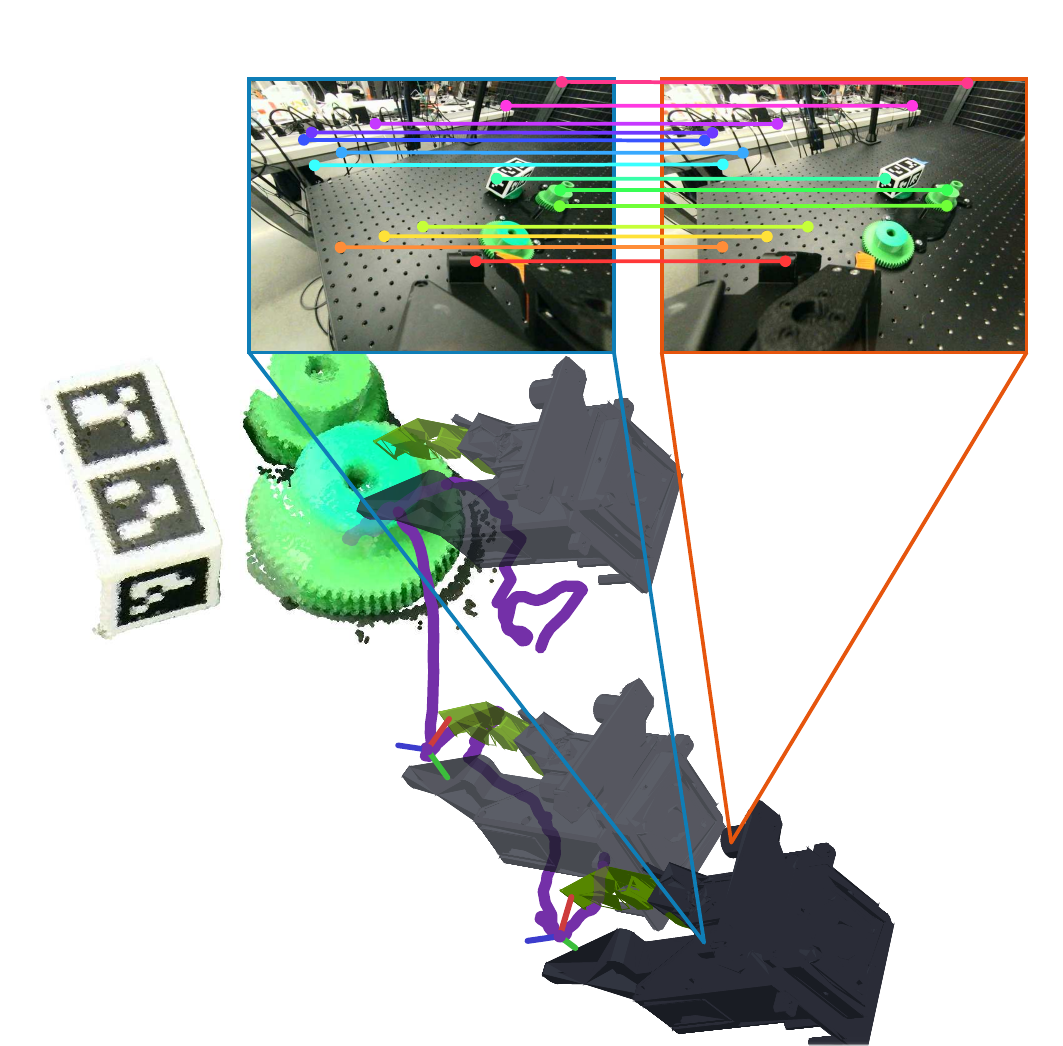}
  \caption{\textbf{Handheld stereo-data collection with \ourmethod.} The purple curve is the recovered end-effector trajectory over the full recording. The insets are the rectified left (blue) and right (orange) frames captured at the starting pose, with SIFT correspondences drawn between them, and the rays run from each lens to its image. The fiducial cube on the left is only for \ourmethod and is not used during data collection for policies. Note that the point cloud, obtained with FoundationStereo \cite{wen2025foundationstereo}, is only used for visualization.}
  \label{fig:stereo_capture}
\end{figure}

Inspired by the above work, we propose a fully open framework with open-source software (OSS) and hardware under a commercially permissive license. In \ourmethod, we simplify cross-embodiment deployment by adopting the ISO 9409-1-50-4-M6 flange mounting standard, allowing the gripper to be mounted directly on common collaborative robots without a new adapter. Our design is modular, built using open electronics modules (e.g., Dynamixel and Arduino). For localization, the original UMI work~\cite{chi2024umi} used monocular-inertial Simultaneous Localization and Mapping (SLAM) with a single GoPro camera. However, a monocular camera, by design, cannot capture metric information. Other recent UMI approaches have proposed using VR tracking devices or native industrial visual-odometry frameworks on smartphones (Apple ARKit)~\cite{xu2025dexumi}, both of which are closed-source and expensive; VR headsets may also require the user to wear a separate device for triangulation. We propose a simpler alternative: using a low-cost stereo camera and running open-source localization frameworks from the SLAM and Structure from Motion (SfM) literature. With multi-view stereo, we obtain metric information for free, as triangulating corresponding pixels across two views (as shown in \figref{fig:stereo_capture}) separated by a known physical baseline resolves absolute scale. Despite the presence of popular SLAM benchmarks for autonomous driving~\cite{geiger2012we} and indoor scene reconstruction~\cite{straub2019replica,handa2014benchmark}, no localization benchmarks exist in the context of policy-based robotic manipulation and UMI. As such, we present a large-scale, stereo-based localization dataset, \benchmark, which allows us to systematically evaluate visual localization in a data distribution representative of UMI-based manipulation policies.

Overall, we make the following contributions with \ourmethod. First, we present a modular, stereo-based open software and hardware framework for sensorized UMI. Second, with this stereo-based UMI gripper, we build a localization benchmark, \benchmark, to evaluate the performance of state-of-the-art, in-the-wild stereo-based localization techniques relative to existing monocular approaches. Third, we release simulation assets for \ourmethod and demonstrate compelling agentic real-to-sim performance on \benchmark, potentially opening paths for future research in sim-to-real and sim-real co-training~\cite{cheng2025generalizable}. Finally, we demonstrate that this system can be used to train action-chunking policies and deploy them on a real robot system.

\section{Related Work}
\label{sec:related_work}

\textbf{Universal Manipulation Interface:} Leader-follower teleoperation rigs~\cite{zhao2023aloha, wu2024gello} collect demonstrations by mapping a kinematically matched leader arm directly to the follower robot. UMI~\cite{chi2024umi} removes the robot from the loop entirely with a trigger-based (but not sensorized) handheld parallel-jaw gripper, pairing a wrist-mounted fisheye camera with a relative end-effector action representation to produce arm-agnostic policies. Several follow-up systems have extended the UMI paradigm along different axes: FastUMI~\cite{zhaxizhuoma2025fastumi} decouples the handheld device from any specific robot gripper and released a large scale dataset over 10k trajectories. HiFi-UMI~\cite{simpleai2026hifiumi} combines head-mounted offline stereo-inertial SLAM to reach millimeter-scale end-effector accuracy. Similar to our work, LEGATO~\cite{seo2024legato} uses the same actuated gripper for both human demonstrations and robot deployment. The gripper is not rigidly attached, but grasped as a tool by the target embodiment gripper. Both Grabette~\cite{pollen2025grabette} and Koala~\cite{hajjahmad2026koala} present a data-capture device together with a purpose-built robot gripper, featuring a separate tracking stereo camera module. \ourmethod similarly develops the handheld device and robot-side gripper as a unified system, but differs in three respects: it uses a modular stereo wrist camera rather than a monocular one used in prior UMI frameworks; its robot-side gripper exposes a standardized ISO~9409-1-50-4-M6 adapter for cross-embodiment deployment; and the entire hardware and software stack is open source. This allows us to use the latest in SLAM and SfM for visual localization, and we present a localization \benchmark dataset using our gripper. \tabref{tab:umi_comparison} further explains \ourmethod in comparison to existing UMI frameworks.

\textbf{Localization:} Prior UMI systems use monocular-inertial SLAM~\cite{young2021visual,song2020grasping,praveena2019characterizing,chi2024umi} to extract the action labels. Its limitation include scale ambiguity, specialized sensor or infrastructure requirements, and sensitivity to motion blur or specilized pre-processing steps to utilize the IMU to get the metric information. Subsequent systems instead use commercial tracking modules, headsets, or external base stations~\cite{zhaxizhuoma2025fastumi,activeumi2025,rdt22026}. Given the promise of stereo over monocular SLAM to get accurate metric reconstruction, \benchmark evaluates how stereo localization affects UMI action-label quality in the wild. SfM estimates camera poses and sparse geometry, after which multi-view stereo (MVS) densifies the reconstruction. COLMAP is a popular pipeline~\cite{schoenberger2016sfm,schoenberger2016mvs}, alongside semi-global stereo matching~\cite{furukawa2010pmvs,hirschmuller2008sgm}. UMI obtains \emph{implicit stereo} using side mirrors~\cite{chi2024umi} but this is only used for policy learning and not localization. \ourmethod instead uses an explicit, calibrated stereo pair to produce metrically grounded action labels from low-cost hardware.

\textbf{SLAM Benchmarks:} SLAM evaluation is organized around domain-specific benchmarks~\cite{macario2022comprehensive}. KITTI established large-scale outdoor driving tasks for visual odometry~\cite{geiger2012we}, while TUM RGBD~\cite{sturm2012benchmark} and ICL-NUIM~\cite{handa2014benchmark} target indoor scene reconstructions. Replica provides high-fidelity, photorealistic indoor scenes for reconstruction~\cite{straub2019replica}. In the context of robotic manipulation, HOPE proposed a benchmark to evaluate 6-DoF household-object pose estimation~\cite{tyree2022hope} while SLAM\&Render~\cite{cerezo2025slamrender} compared SLAM with neural rendering for tabletop scenes. However, to the best of our knowledge, we present the first SLAM benchmark (\benchmark) with policy action labels, specifically to quantitatively study localization for UMI-based manipulation.

\section{\ourmethod}
\label{sec:method}

With \ourmethod, we present a modular, stereo-based framework for UMI, as shown in \figref{fig:teaser}. \tabref{tab:umi_comparison} situates \ourmethod relative to other open UMI frameworks.

\begin{table*}[t]
  \centering
  \caption{\textbf{Comparing \ourmethod with other UMI frameworks.} BOM costs are approximate totals in USD (\ourmethod is itemized in Appendix~\ref{sec:appendix_bom}); BOM = Bill of Materials, OSS = Open Source Software.}
  \label{tab:umi_comparison}
  \footnotesize
  \begin{tabularx}{\textwidth}{l c c c c X X}
    \toprule
    Method & Approximate BOM Cost (USD) & Weight (kg) & Sim & Deployment Gripper & Mounting Adapter & Localization \\
    \midrule
    UMI~\cite{chi2024umi} & 5,389 (includes WSG-50 Gripper) & 0.650 & \nocell & \nocell & Finger Mounts & OSS Monocular-IMU SLAM \\
    FastUMI Pro~\cite{zhaxizhuoma2025fastumi} & 840 & -- & \nocell & \nocell & Finger Mounts & Specialized SLAM Camera \\
    LEGATO~\cite{seo2024legato} & 1{,}218 & -- & \yescell & \yescell & Gripper grasps tool (not rigidly attached) & Specialized SLAM Camera \\
    DexUMI~\cite{xu2025dexumi} & 1,316 (includes iPhone) & -- & \nocell & \nocell & -- & iPhone ARKit \\
    HandUMI & 454 (includes VR Tracker) & -- & \nocell & \nocell & Finger Mounts & VR Tracker \\
    Grabette~\cite{pollen2025grabette} & 704 & -- & \nocell & \yescell & -- & OSS Stereo + IMU SLAM \\
    \textbf{\ourmethod (ours)} & \textbf{$\sim$400} & \textbf{0.408} & \textbf{\yescell} & \textbf{\yescell} & \textbf{Standardized ISO} & \textbf{OSS Stereo SLAM/SfM} \\
    \bottomrule
  \end{tabularx}
\end{table*}

\subsection{Open-Source Modular UMI Gripper}
\label{subsec:gripper}

The original UMI~\cite{chi2024umi} established the paradigm of collecting robot-free motion data. We build upon this work with the goal of further simplifying deployment. Our design principles are as follows:

\begin{itemize}
  \item \textbf{Open modules.} We build the system exclusively from open, off-the-shelf electronics: a Dynamixel XC430-W150-T actuator, an Arduino OpenRB 150 microcontroller, an Arduino MKR IMU shield, and a commodity stereo camera. All structural parts are 3D printed. We deliberately avoided designing a custom printed circuit board (PCB). Although a PCB would result in a more streamlined design, it would complicate widespread adoption of the gripper, since a bespoke PCB would need to be fabricated before anyone could build the device. With the current design, users can simply purchase the components in the bill of materials (BOM) for direct assembly.
  \item \textbf{Modular.} The camera and fingers are interchangeable as long as they follow the mounting specifications. We follow the finger-mounting pattern from~\cite{chi2024umi}.
  \item \textbf{Seamless system integration.} We standardize on the ISO~9409-1-50-4-M6 mount, as many collaborative robot arms have coalesced around this standard. For robots that do not provide this mount, such as the YAM robot, a custom adapter can bridge the robot interface to the ISO mount, allowing \ourmethod to be attached.
  \item \textbf{Low cost.} By default, we use a low-cost global-shutter fisheye stereo camera together with a Dynamixel motor, though either can be replaced with an alternative. Unlike a rolling-shutter camera, a global-shutter camera does not skew the image under the fast handheld motion typical of in-the-wild collection, thereby preserving the feature correspondences on which localization depends.
\end{itemize}

\subsubsection*{Gripper Mechanism and Design}

We adopt a simple two-finger revolute mechanism for the gripper. To reduce complexity and gripper weight, only one side is actuated, though the passive side can readily be swapped for a second actuated one. To support diverse pick-and-place tasks in the wild, we adopt the finger design from Grasp EveryThing (GET)~\cite{Burgess2025}, omitting its tactile sensing hardware. Our custom deformable fingers, printed in TPU, provide natural compliance without compromising grasp strength.

\subsection{Stereo-Localization}
\label{subsec:stereo_localization}

Our localization builds on a stereo camera and classical triangulation, which determines the 3D position of a scene point by intersecting the light rays projected through two calibrated camera centers. An example of a data collection session is shown in \figref{fig:stereo_capture}. We run a camera calibration routine to recover the per-camera intrinsics and the stereo baseline, which we later show to be a key hyperparameter in our ablation studies. Because we use fisheye lenses, we first undistort the images and then rectify them; localization is run on the resulting rectified pair. We use SAM~\cite{carion2025sam3} to obtain segmentation masks for the gripper and the fiducials in the benchmark dataset, and instruct the localization algorithms to ignore features within the masked regions. We find this approach to be more stable because the gripper fingers move and may affect feature matching.

\section{Evaluating UMI Localization: \benchmark}
\label{sec:benchmark}

\begin{figure*}[t]
  \centering
  \input{figures/figure3/figure3_seq}
  \label{fig:rollouts}
\end{figure*}

We evaluate and benchmark the localization performance of SLAM algorithms for in-the-wild settings.

\subsubsection*{Data Capture Protocol}

To obtain ground truth, we use an AprilTag cube fitted with ArUco markers. For a fair comparison, we do not want the SLAM method to exploit the markers themselves as tracking features, so the cube is masked out using a segmentation mask.

\subsubsection*{Tasks}

We select a wide variety of tasks, spanning several forms of pick-and-place, kitting, and long-horizon non-prehensile manipulation, across different environments. Example rollouts are shown in \figref{fig:rollouts}. Further task and protocol details are provided in Appendix~\ref{sec:appendix_experiments}.

\subsubsection*{Quantitative Metrics}

We report three quantities; see Appendix~\ref{sec:appendix_metrics} for their formal definitions.

\textbf{Translational Absolute Trajectory Error (translational ATE).} The estimated trajectory is aligned to ground truth with a scale-fixed Umeyama SE(3) alignment~\cite{umeyama1991}, after which we take the per-frame translation residual norm and report it as RMSE in meters, shown as millimeters in the tables, alongside the mean, median, standard deviation, minimum, and maximum. This is the standard TUM RGB-D ATE~\cite{sturm2012benchmark}.

\textbf{Rotational Absolute Trajectory Error (rotational ATE).} Rotational ATE is computed from the same aligned per-frame SE(3) residuals as translational ATE, but uses the geodesic rotation angle and reports its RMSE in degrees. A constant sensor-to-ground-truth frame-convention offset, taken as the chordal mean of the residual rotations, is subtracted before scoring. This metric is not part of the original TUM benchmark and is our rotational analogue of ATE.

\textbf{Runtime.} Runtime is the wall-clock time (in seconds) around each localization run. For UMI-SLAM, this includes the one-time mapping cost for that scene.

\section{Results and Discussion}
\label{sec:results}

In our experiments, we seek to answer the following questions: (1) Do state-of-the-art stereo-based localization methods outperform monocular approaches on \benchmark? (2) What are the key hyperparameters influencing our stereo-based localization approach? (3) Is the \ourmethod gripper sufficiently robust for real-robot policy learning? (4) Can \ourmethod support real-to-sim reconstruction and successful demonstration playback in a digital twin?

\subsection{Localization Evaluations on UMI-Loc-Bench}
\label{subsec:results_benchmark}

\begin{figure*}[t]
  \centering
  \begin{subfigure}[b]{0.47\textwidth}
    \includegraphics[width=\linewidth]{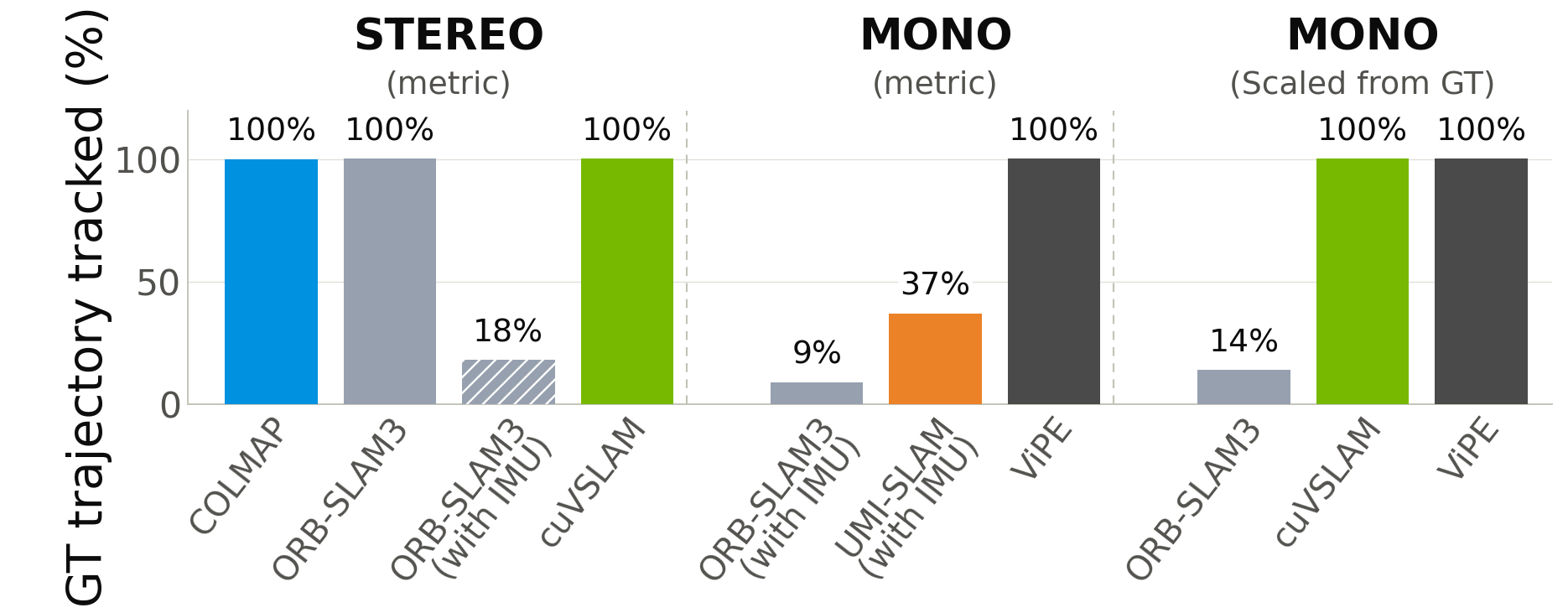}
    \caption{Full-dataset trajectory coverage}
    \label{fig:bench_coverage}
  \end{subfigure}
  \hfill
  \begin{subfigure}[b]{0.47\textwidth}
    \includegraphics[width=\linewidth]{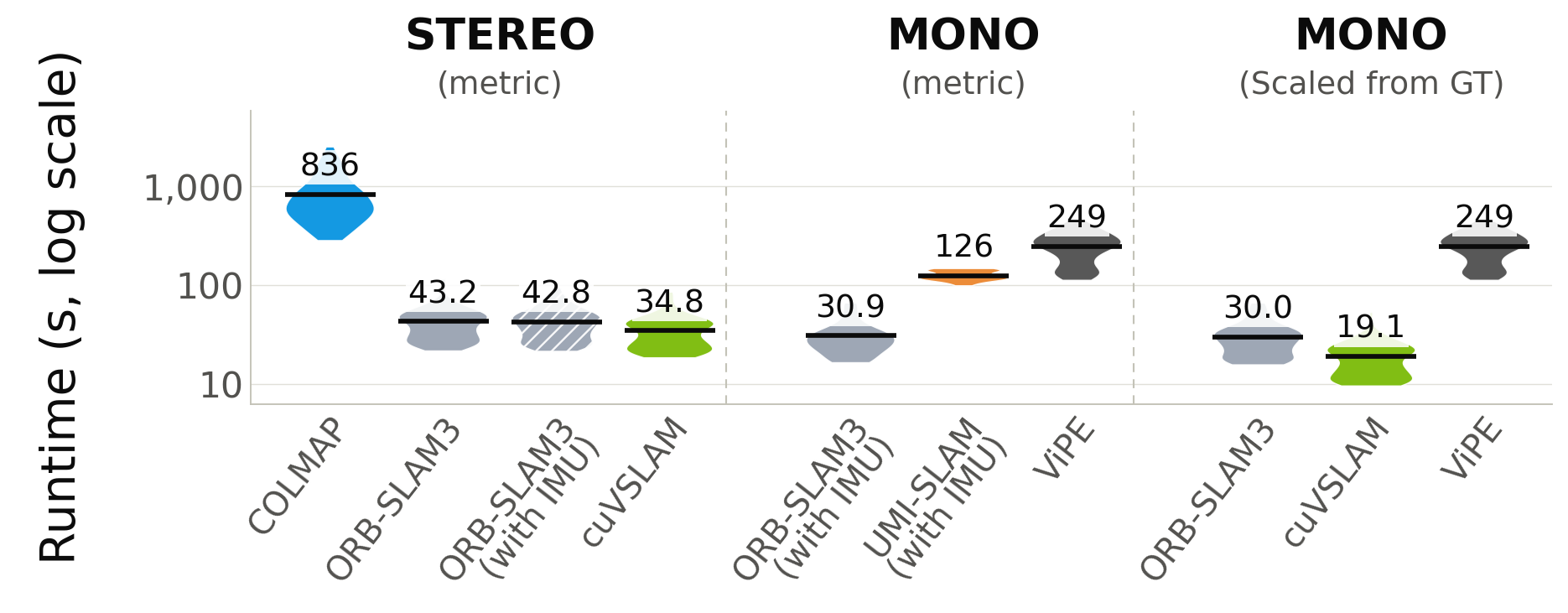}
    \caption{Full-dataset runtime}
    \label{fig:bench_runtime}
  \end{subfigure}
  \\
  \begin{subfigure}[b]{0.47\textwidth}
    \includegraphics[width=\linewidth]{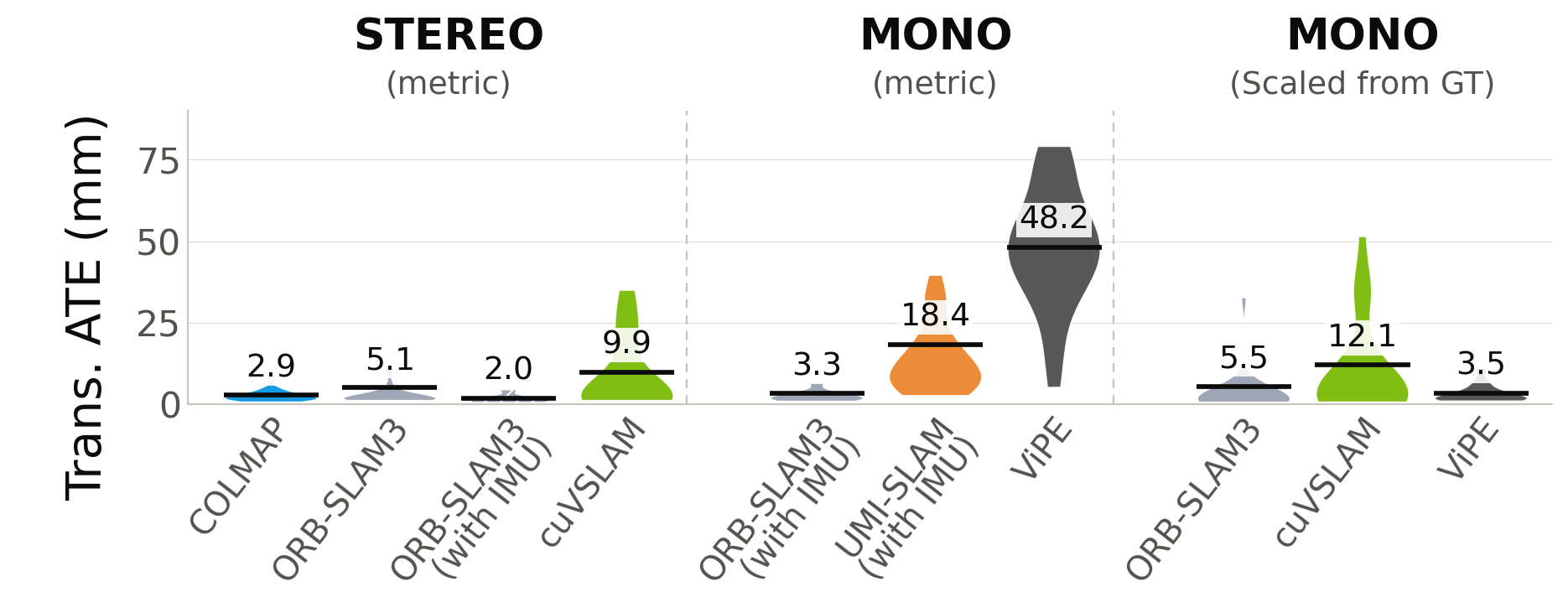}
    \caption{UMI-SLAM common-segment translational ATE}
    \label{fig:bench_tracked_ate_trans}
  \end{subfigure}
  \hfill
  \begin{subfigure}[b]{0.47\textwidth}
    \includegraphics[width=\linewidth]{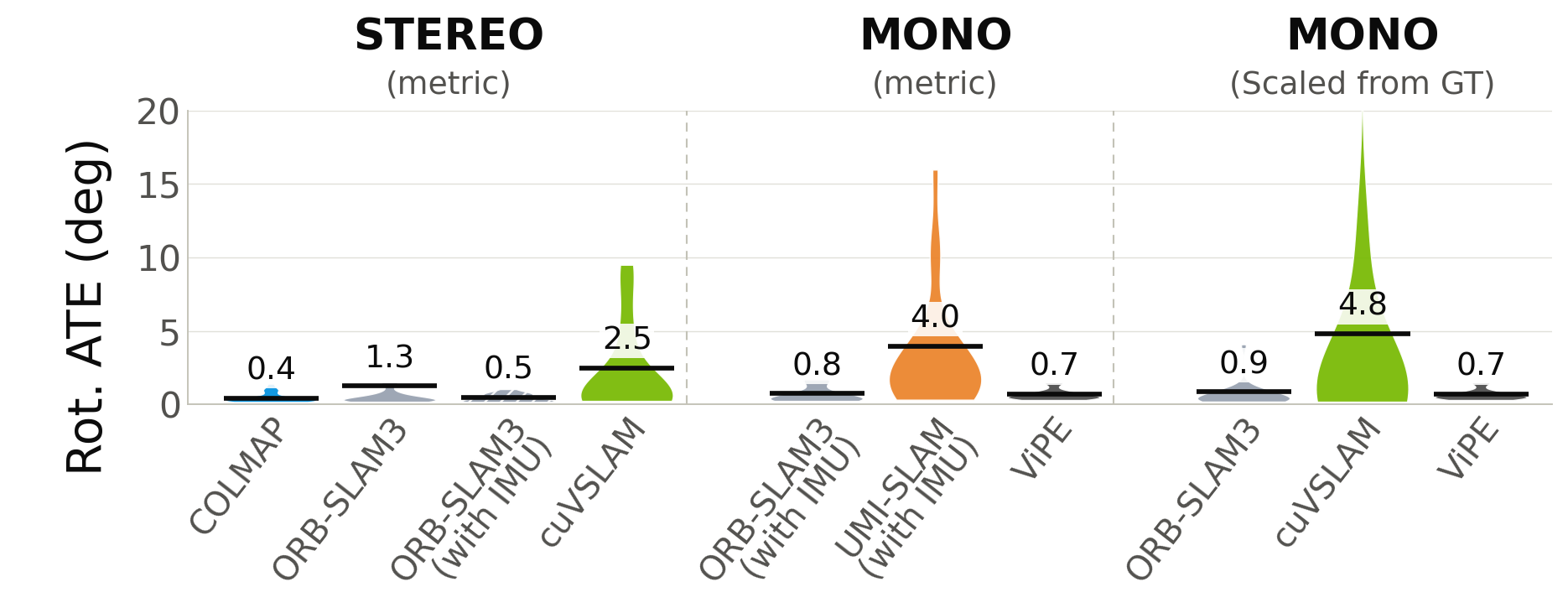}
    \caption{UMI-SLAM common-segment rotational ATE}
    \label{fig:bench_tracked_ate_rot}
  \end{subfigure}
  \\
  \begin{subfigure}[b]{0.47\textwidth}
    \includegraphics[width=\linewidth]{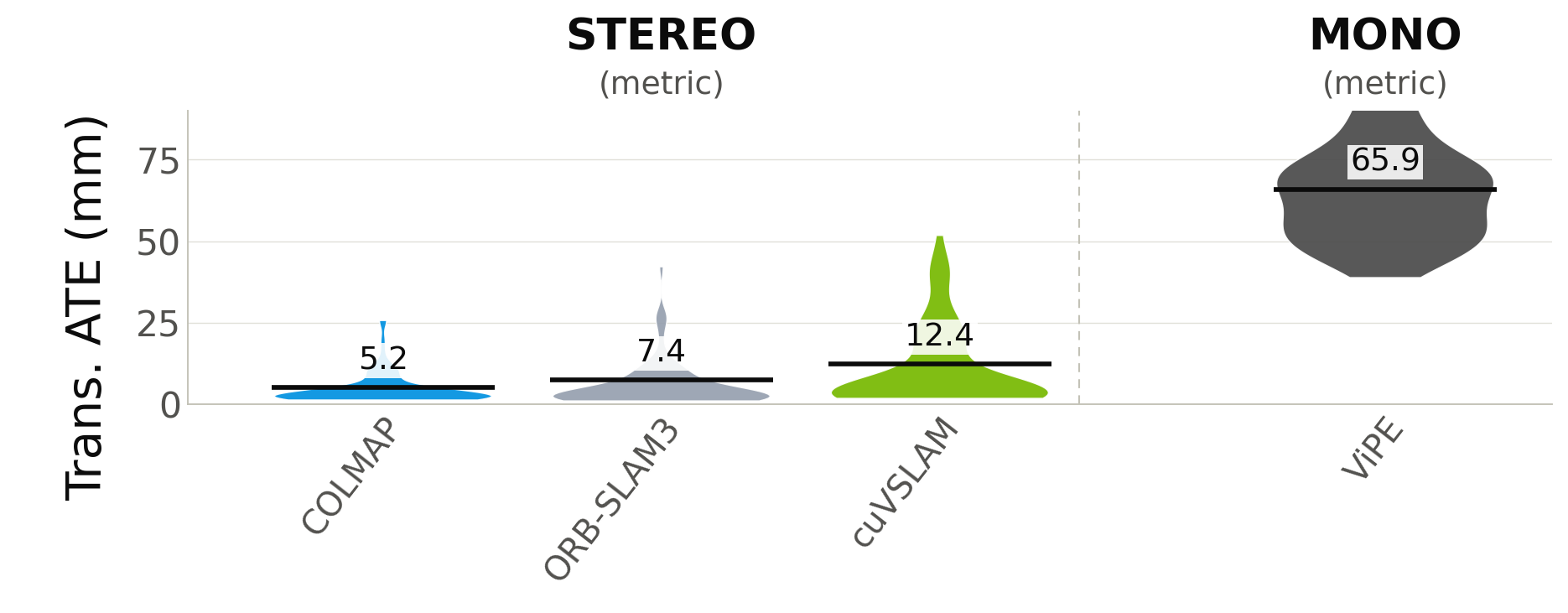}
    \caption{Full-trajectory translational ATE}
    \label{fig:bench_stereo_ate_trans}
  \end{subfigure}
  \hfill
  \begin{subfigure}[b]{0.47\textwidth}
    \includegraphics[width=\linewidth]{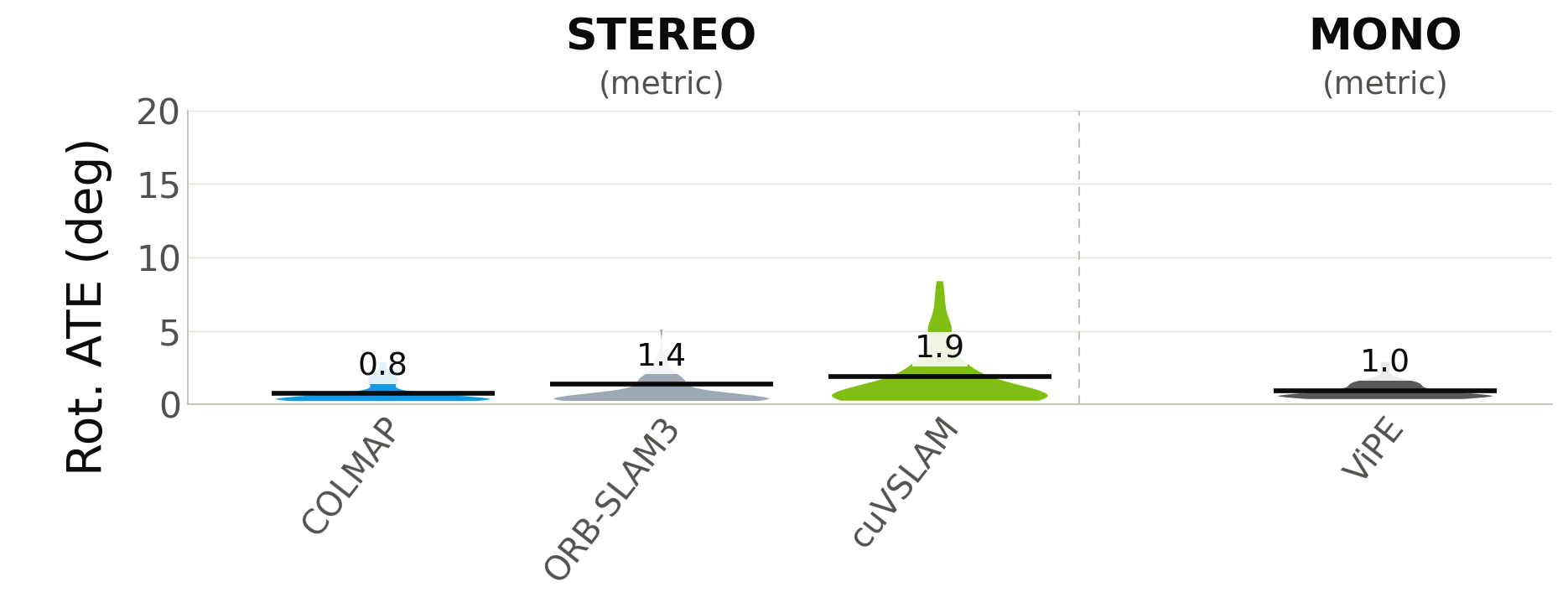}
    \caption{Full-trajectory rotational ATE}
    \label{fig:bench_stereo_ate_rot}
  \end{subfigure}
  \caption{\textbf{Localization results on \benchmark.} (a)--(b) Coverage and runtime over the benchmark demonstrations. (c)--(d) ATE on UMI-SLAM-tracked ground-truth segments (27 demonstrations); all methods use identical segments. (e)--(f) Full-trajectory ATE for stereo COLMAP~\cite{schoenberger2016sfm}, ORB-SLAM3~\cite{campos2021orbslam3}, and cuVSLAM~\cite{korovko2025cuvslam}, compared with native-scale monocular ViPE~\cite{huang2025vipe}. Panels (a)--(d) evaluate the same ViPE run at native scale and version scaled based on the ground truth; its two runtime bars therefore denote one run. UMI-SLAM uses the original implementation~\cite{chi2024umi}. Black rules mark means; violins show the central 2--98\% and are clipped at the displayed limits.}
  \label{fig:bench_results}
\end{figure*}

\textbf{Experimental Details.} We evaluate representative SLAM and SfM backends spanning stereo, monocular, and monocular-inertial settings. We run COLMAP~\cite{schoenberger2016sfm} as a stereo-rig reconstruction (\S\ref{subsec:stereo_localization}); it does not ingest IMU measurements. cuVSLAM~\cite{korovko2025cuvslam} is a real-time, CUDA-accelerated system supporting stereo, stereo-inertial, and monocular inputs, while ORB-SLAM3~\cite{campos2021orbslam3} is the feature-based backend adopted by UMI~\cite{chi2024umi}. While classical SLAM/SfM can be proficient in estimating camera poses in long sequences, they can be less robust to dynamic objects or poorly textured environments. As such, we also evaluate ViPE~\cite{huang2025vipe}, a SOTA learned monocular video-pose model with near-metric output. UMI's modified ORB-SLAM3 mono+IMU pipeline was introduced because the original heuristic initialization was fragile and lengthy under fast motion; we call this baseline UMI-SLAM. It loads a prerecorded map before resuming normal SLAM and uses a known-size fiducial marker during mapping to stabilize initialization; the marker is absent during demonstration collection. We use the original implementation and marker, with one mapping demonstration for each of the nine tasks. Although a fiducial marker provides ground truth, we mask it during evaluation to test in-the-wild localization without marker-based tracking. For the monocular baselines, they all use the left stereo camera image.

\textbf{Stereo vs. Monocular.} \figref{fig:bench_coverage}, \figref{fig:bench_tracked_ate_trans}, and \figref{fig:bench_tracked_ate_rot} compare stereo and monocular localization. A calibrated stereo pair directly observes depth and recovers metric trajectories from vision alone; a monocular camera cannot recover absolute scale without an additional source such as an IMU. Accordingly, the rightmost monocular group is scale-aligned to ground truth to isolate trajectory geometry and is not directly comparable to natively metric estimates. All stereo methods track essentially the full trajectory, whereas ORB-SLAM3 mono+IMU and UMI-SLAM average only \SI{9}{\percent} and \SI{37}{\percent} coverage. ViPE completes 89 of 90 demonstrations and covers essentially every frame of those completed runs, but its native-scale common-segment translational ATE is \SI{48.2}{\milli\meter}; fitting one scale to ground truth reduces it to \SI{3.5}{\milli\meter}, while rotational ATE remains \SI{0.7}{\degree}. Thus, ViPE recovers strong monocular trajectory geometry, but not the metric scale required for action labels. For a fair accuracy comparison, \figref{fig:bench_tracked_ate_trans} and \figref{fig:bench_tracked_ate_rot} evaluate every method only on the corresponding ground-truth segments tracked by UMI-SLAM. On these segments, COLMAP achieves \SI{2.9}{\milli\meter}/\SI{0.4}{\degree}, versus \SI{18.4}{\milli\meter}/\SI{4.0}{\degree} for UMI-SLAM; scale-fitted cuVSLAM mono remains worse than cuVSLAM stereo (\SI{12.1}{\milli\meter}/\SI{4.8}{\degree} versus \SI{9.9}{\milli\meter}/\SI{2.5}{\degree}).

\textbf{Stereo Comparisons.} \figref{fig:bench_stereo_ate_trans} and \figref{fig:bench_stereo_ate_rot} report full-trajectory ATE. Among the pure-stereo methods, COLMAP is most accurate at \SI{5.2}{\milli\meter}/\SI{0.8}{\degree}, compared with \SI{7.4}{\milli\meter}/\SI{1.4}{\degree} for ORB-SLAM3 and \SI{12.4}{\milli\meter}/\SI{1.9}{\degree} for cuVSLAM. Native-scale ViPE records \SI{65.9}{\milli\meter}/\SI{1.0}{\degree}; for reference, fitting its scale to ground truth reduces translational ATE to \SI{6.0}{\milli\meter} (median fitted scale: \num{0.556}). Runtime is \SI{836}{\second} for COLMAP, \SI{249}{\second} for ViPE, \SI{43.2}{\second} for ORB-SLAM3, and \SI{34.8}{\second} for cuVSLAM (\figref{fig:bench_runtime}). COLMAP's offline global bundle adjustment explains its higher runtime cost, while cuVSLAM offers a useful accuracy--throughput tradeoff for processing handheld demonstrations at scale. We use cuVSLAM for our policy learning experiments.

\textbf{Classical vs. Neural Localization.} We evaluate ViPE~\cite{huang2025vipe}, a SOTA learned neural localization method. From \figref{fig:bench_tracked_ate_trans}, we found that ViPE actually successfully recovered a trajectory that was geometrically consistent with the ground truth. If we use the ground truth scale, it has a low ATE error (\SI{3.5}{\milli\meter}/\SI{0.7}{\degree} comparable with the stereo methods. Unfortunately, it had a poor prediction of the depth and camera intrinsics, resulting in an inaccurate metric reconstruction. This was mostly owing to the fact that ViPE was trained on outdoor or room-scale datasets, as such this is out-of-distribution for the wrist-mounted UMI setup. Future work can consider finetuning ViPE for \ourmethod.

\textbf{Usefulness of IMU.} COLMAP does not consume IMU data, so we isolate the effect of the IMU using ORB-SLAM3, which supports both stereo and stereo-inertial modes. Adding the IMU reduces mean translational/rotational ATE from \SI{5.1}{\milli\meter}/\SI{1.3}{\degree} to \SI{2.0}{\milli\meter}/\SI{0.5}{\degree} in \figref{fig:bench_tracked_ate_trans}. This improvement, however, does not translate into a reliable overall trajectory: coverage falls from \SI{100}{\percent} to \SI{18}{\percent} (\figref{fig:bench_coverage}), so the stereo-inertial configuration loses tracking over most of each sequence. Overall, these failure modes show that using a IMU incures additional system integration costs and tuning for the small workspace and handheld dynamics of UMI-style manipulation.

\textbf{Importance of Mapping and Marker-based Initialization.} The original UMI work~\cite{chi2024umi} found that the standard ORB-SLAM3 mono+IMU SLAM implementation was inadequate and prone to error in the UMI manipulation setting: its heuristic map, keyframe, and IMU initialization was fragile and lengthy under fast motion, producing no pose estimates and wasting substantial portions of the recording. UMI-SLAM addresses this failure by initializing from a prerecorded map and using a known-size fiducial marker to disambiguate monocular-inertial initialization. We observe the same benefit on the \ourmethod rig: relative to unmodified ORB-SLAM3 mono+IMU, UMI-SLAM improves mean trajectory coverage from \SI{9}{\percent} to \SI{37}{\percent}. Unfortunately, this improvement requires additional, scene-dependent preprocessing. We use the original UMI-SLAM implementation and marker and collect a separate mapping demonstration for each of the nine \benchmark tasks, following the instructions reported in the UMI paper~\cite{chi2024umi}. Mapping is itself an ambiguous and cumbersome data-collection step because the mapped scene configuration may differ from the scene encountered later, especially after manipulation changes object poses or the environment. Despite map-based and marker-enhanced initialization, UMI-SLAM is still less stable and less accurate than the pure-stereo methods. It tracks only \SI{37}{\percent} on average and yields translational/rotational ATE of \SI{18.4}{\milli\meter}/\SI{4.0}{\degree}, whereas the stereo methods track essentially the entire trajectory without requiring a mapping run, fiducial initialization, or other scene-specific preprocessing.

\subsection{Real Policy Deployment}
\label{subsec:results_real_robot}

We demonstrate that \ourmethod can be used for real-world UMI-based policy learning. We show deployments on a UR10 robot for block-stacking and pick-and-place tasks. Our control stack and policy is Diffusion Policy (DP)~\cite{chi2023diffusionpolicy}, matching the original UMI paper, with a relative Cartesian end-effector action space~\cite{chi2024umi}. We emphasize that \ourmethod is agnostic to the choice of downstream policy: DP is used here only to match the original UMI setting, and nothing in \ourmethod's hardware or localization stack precludes swapping in other, more recent chunk-predicting policy classes, such as Vision-Language-Action models (VLAs)~\cite{black2024pi0,team2025geminirobotics,nvidia2025gr00tn1} or World-Action Models (WAMs), which we leave to future work. The policy is trained to predict actions in the end-effector (TCP) frame. We collect 200 demonstrations for block stacking and 50 for pick-and-place, and use cuVSLAM for localization. For pick-and-place, both the pose of the box and the initial pose of the camera are randomized during data collection. At deployment, the gripper attaches directly onto the robot arm, requiring no recalibration between data collection and deployment.

\begin{figure}[t]
  \centering
  \includegraphics[width=0.19\linewidth]{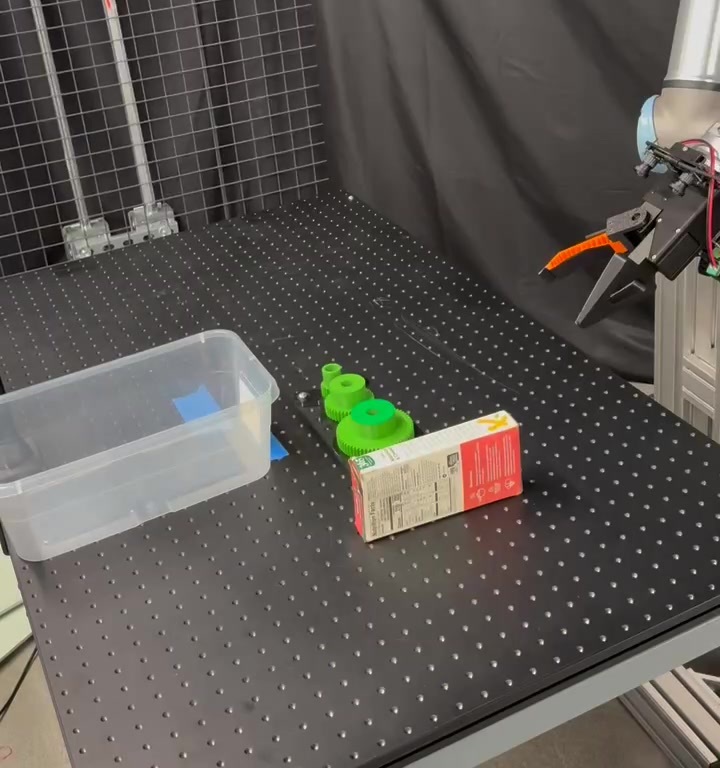}\hfill
  \includegraphics[width=0.19\linewidth]{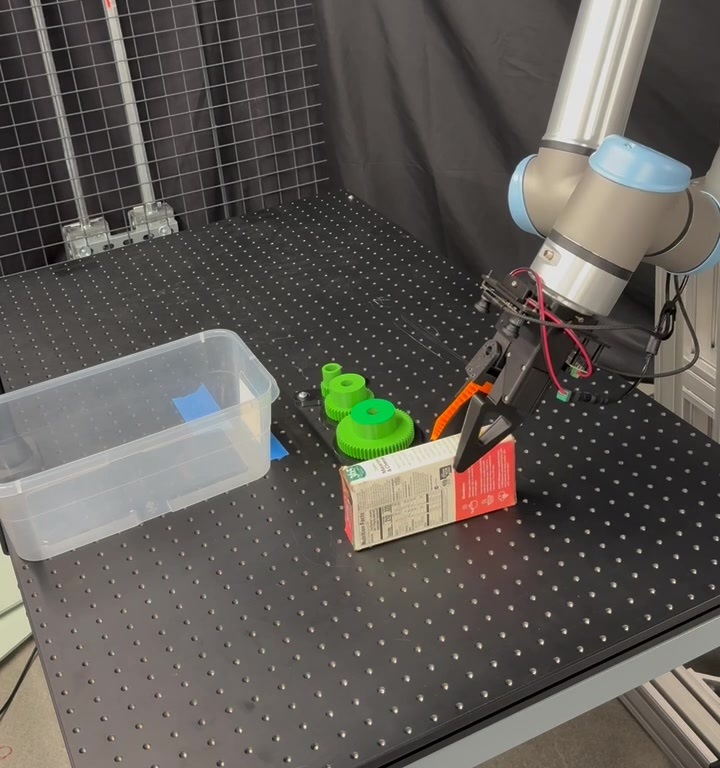}\hfill
  \includegraphics[width=0.19\linewidth]{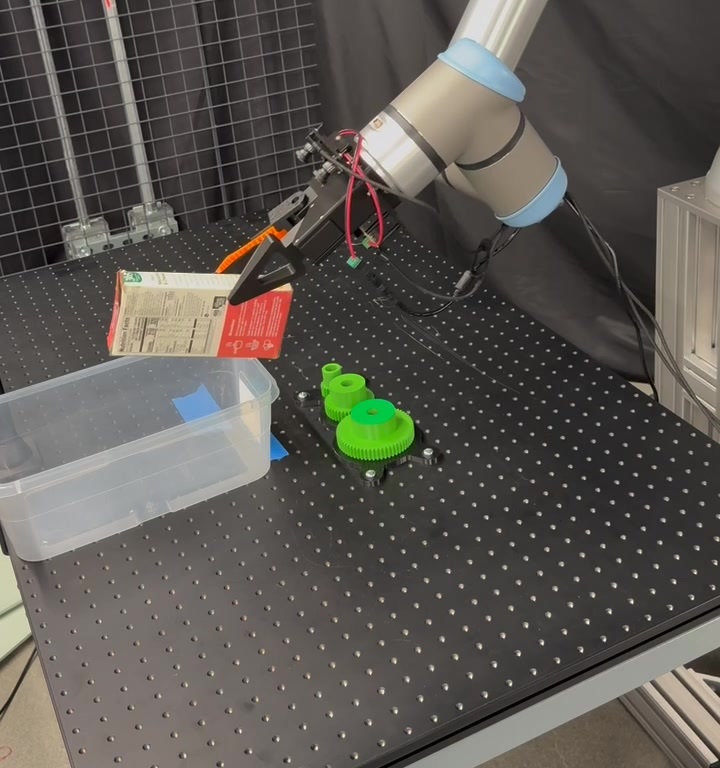}\hfill
  \includegraphics[width=0.19\linewidth]{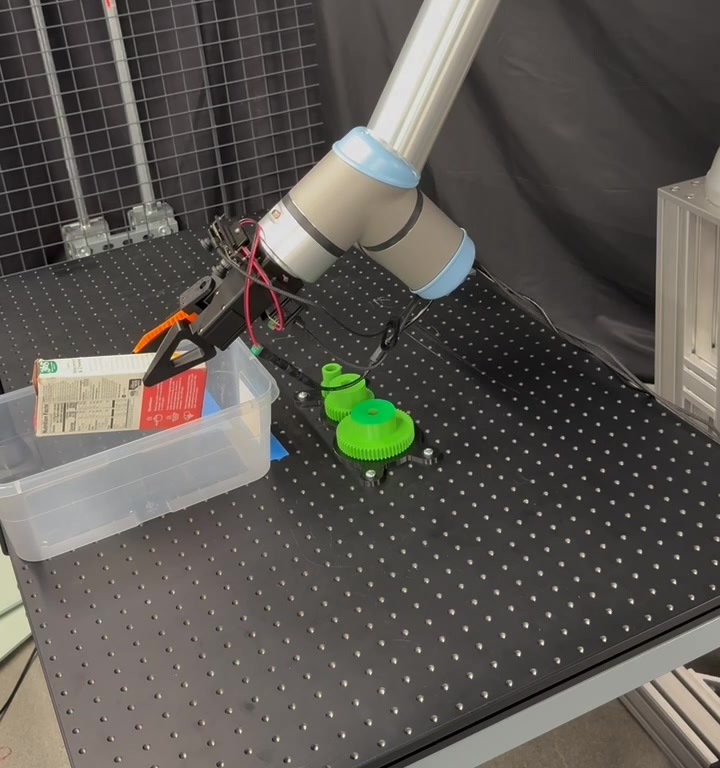}\hfill
  \includegraphics[width=0.19\linewidth]{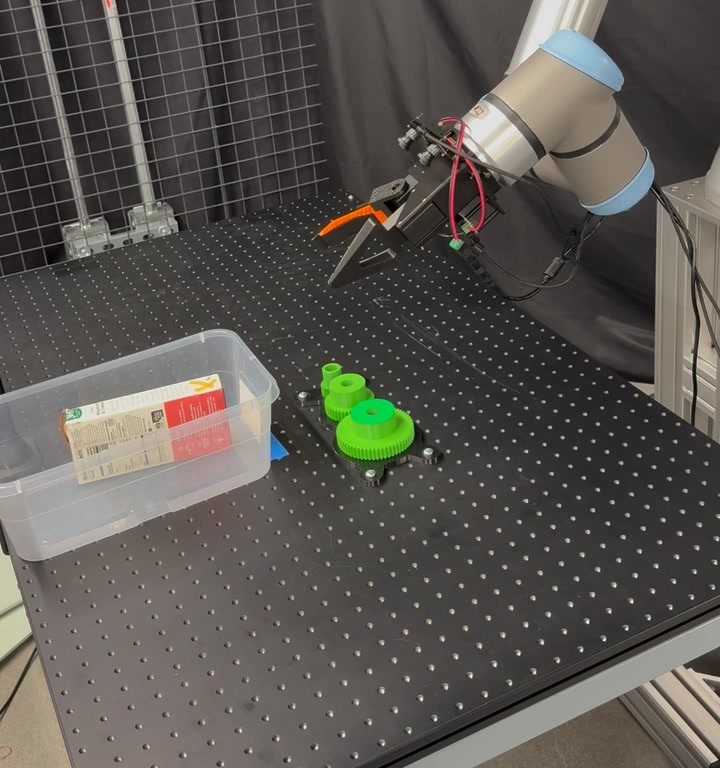}\\[0.5ex]
  \includegraphics[width=0.19\linewidth]{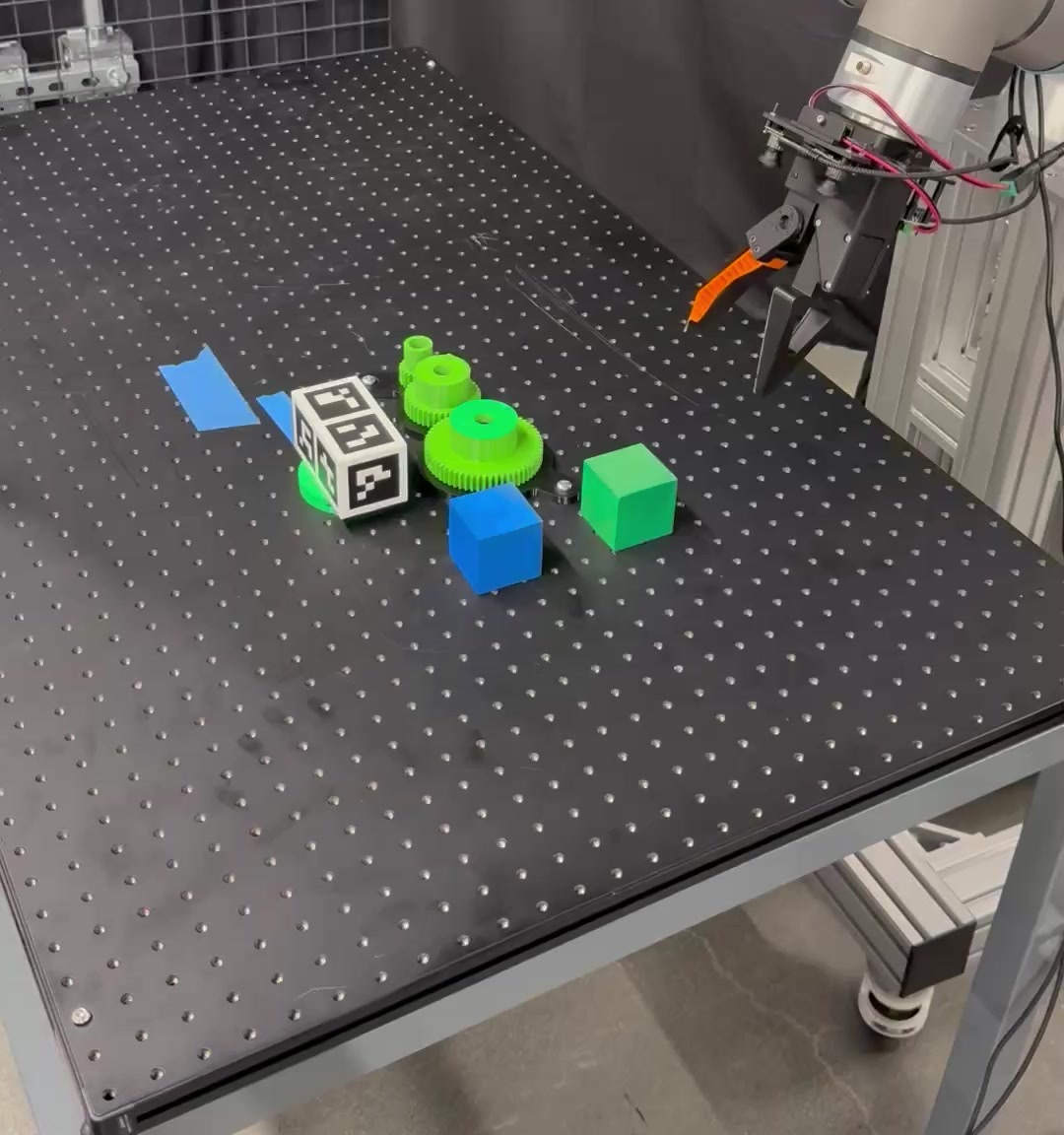}\hfill
  \includegraphics[width=0.19\linewidth]{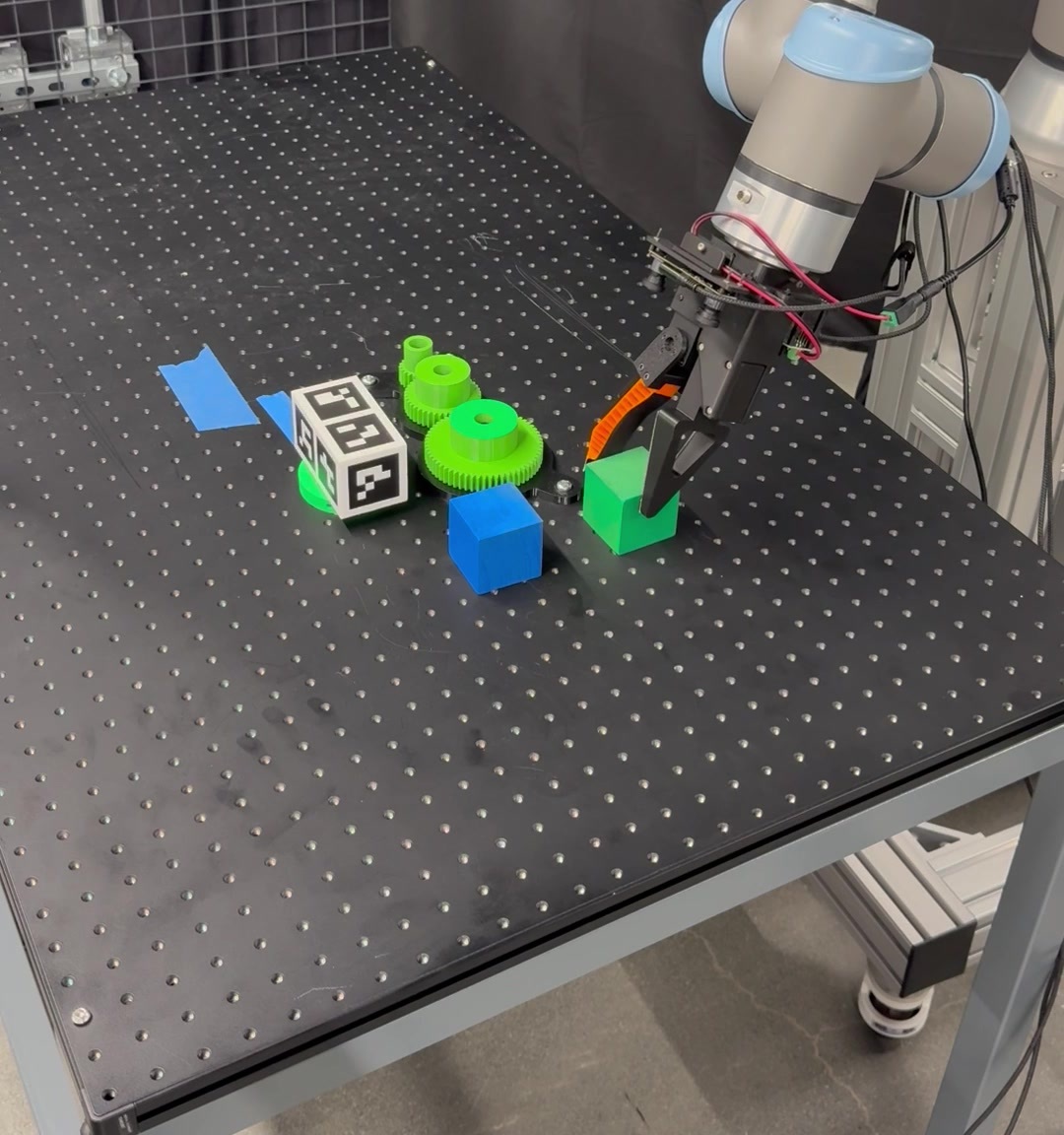}\hfill
  \includegraphics[width=0.19\linewidth]{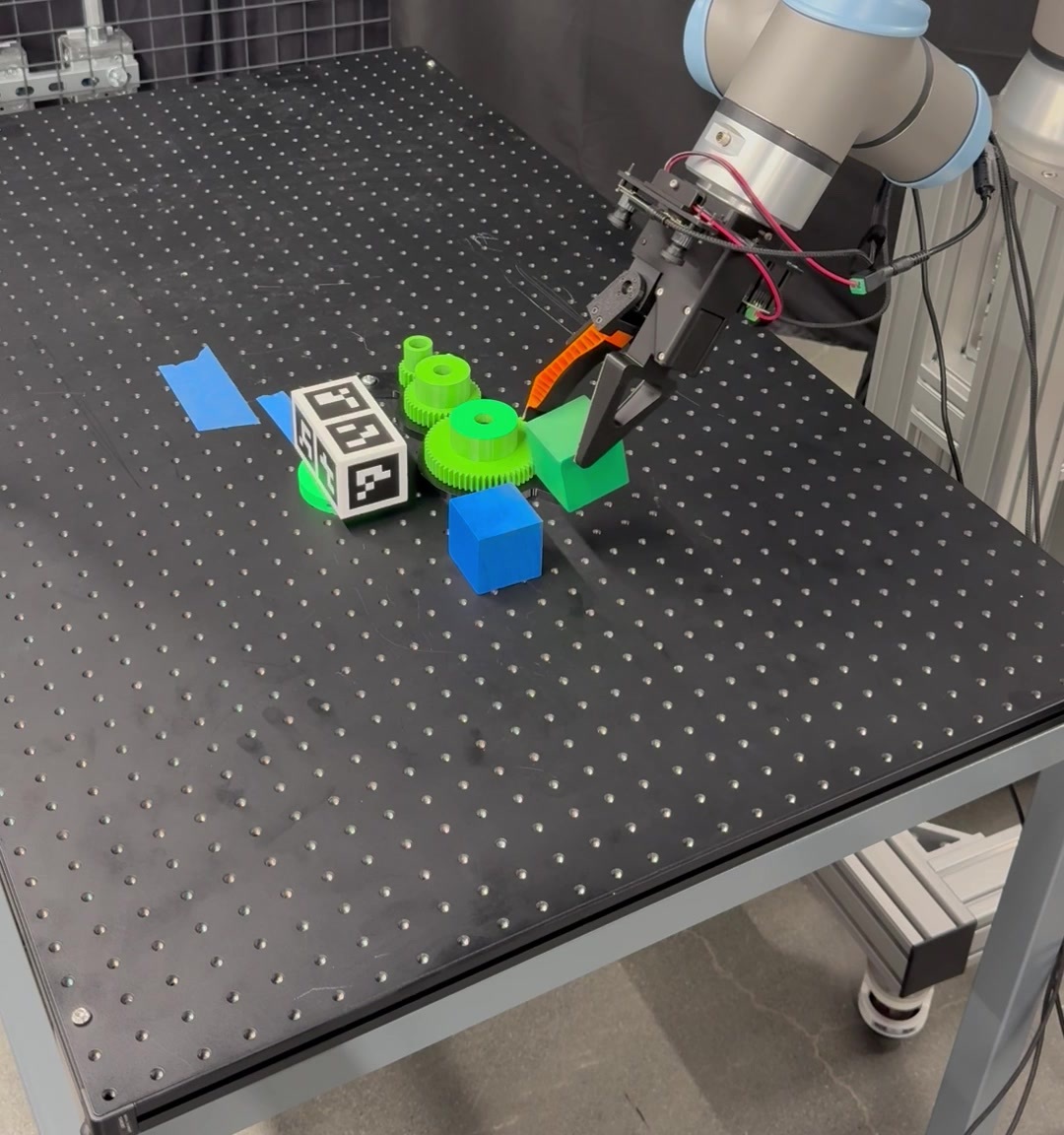}\hfill
  \includegraphics[width=0.19\linewidth]{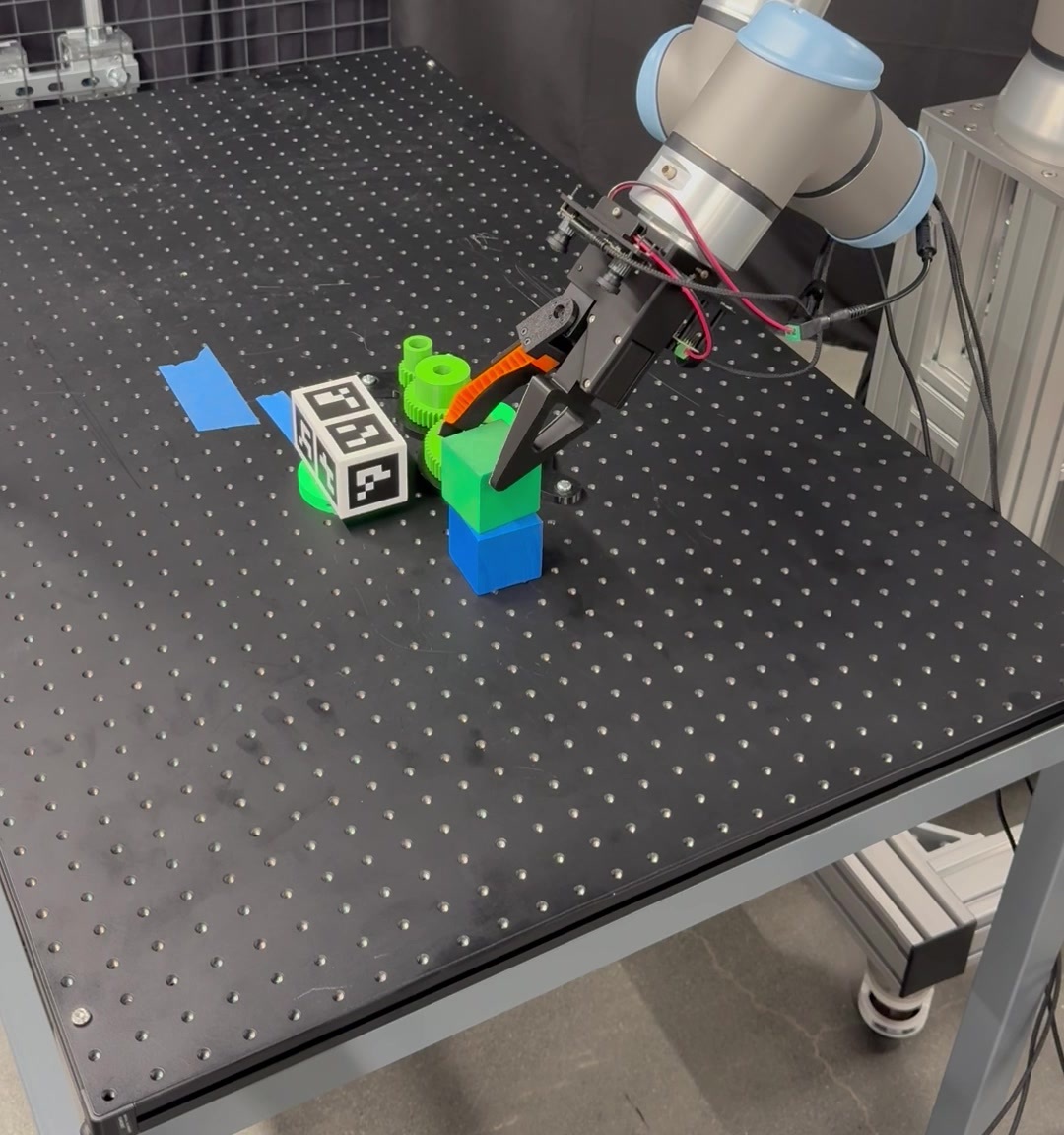}\hfill
  \includegraphics[width=0.19\linewidth]{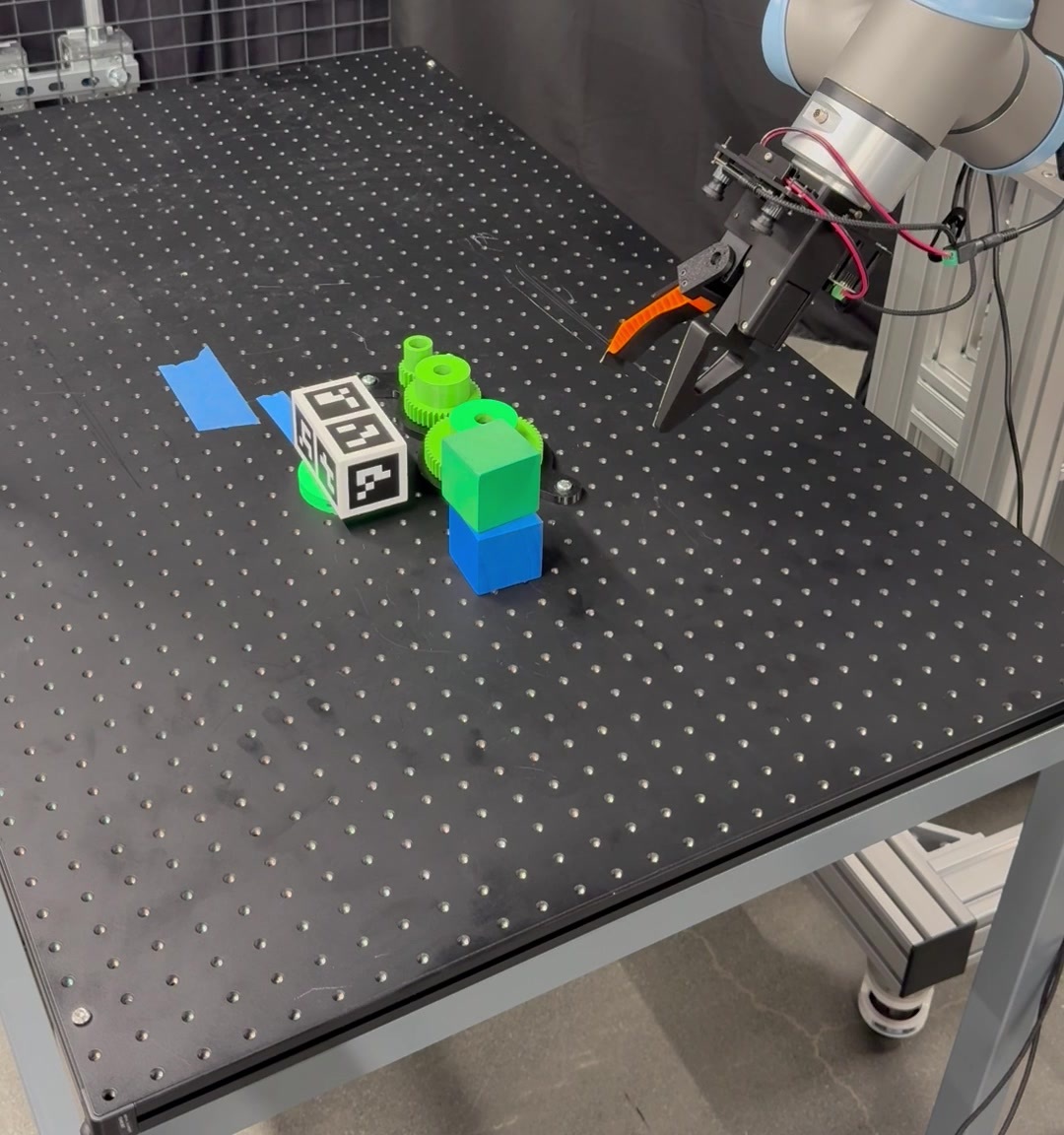}
  \caption{\textbf{Real robot rollouts.} Representative UR10 policy rollouts for pick-and-place \figtop and block stacking \figbottom.}
  \label{fig:real_pnp_rollout}
\end{figure}

Overall, our policies achieve a \SI{90}{\percent} success rate on both tasks evaluated for 10 trials under varied initial conditions. \figref{fig:real_pnp_rollout} shows representative policy rollouts for both tasks. For pick-and-place, a trial is deemed successful only if the target is picked and placed inside the bin without the object colliding with the bin. The one failure for pick-and-place was due to grasp slippage. In the block-stacking task, the policy dropped the block once before placing it on the bottom one.

\subsection{Ablation Experiments}
\label{subsec:results_ablations}

\textbf{Sensitivity to Baseline Calibration.} Stereo localization obtains metric scale from the calibrated camera baseline, and any inaccuracies in it affect the final translational error.

\begin{wrapfigure}{r}{0.50\columnwidth}
  \vspace{-0.6\baselineskip}
  \centering
  \includegraphics[width=\linewidth]{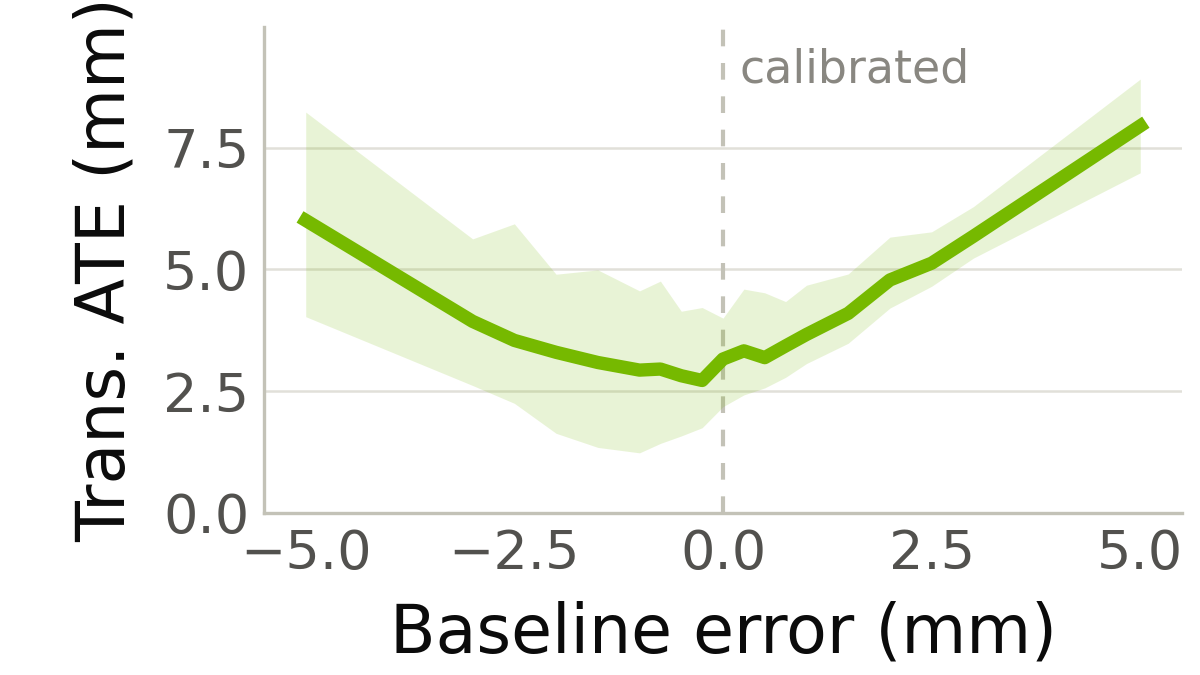}
  \caption{cuVSLAM stereo~\cite{korovko2025cuvslam} under synthetic calibrated-baseline perturbations across several demonstrations. Solid: mean translational ATE; shaded: min--max; dashed: calibrated baseline.}
  \label{fig:ablation_baseline}
  \vspace{-0.5\baselineskip}
\end{wrapfigure}
\noindent We empirically test this by synthetically perturbing the \SI{59.7}{\milli\meter} baseline supplied to cuVSLAM from \SI{-5}{\milli\meter} to \SI{+5}{\milli\meter} without re-rectifying the images, using results averaged over the \benchmark dataset. In \figref{fig:ablation_baseline}, mean translational ATE is \SI{3}{\milli\meter} at the calibrated baseline and reaches its minimum of \SI{2.7}{\milli\meter} at a \SI{-0.25}{\milli\meter} perturbation, indicating that the calibration is close to optimal within the resolution of this sweep. The error rises to \SI{6}{\milli\meter} and \SI{8.0}{\milli\meter} at \SI{-5}{\milli\meter} and \SI{+5}{\milli\meter}, respectively, while rotational ATE varies by only \SI{0.1}{\degree} across the entire range, confirming that the error in the baseline mainly affects the trajectory scale. We obtain $\sim$0.18\,pixel reprojection error independently for each eye.

\begin{wrapfigure}{r}{0.49\columnwidth}
  \vspace{-1.0\baselineskip}
  \centering
  \includegraphics[width=\linewidth]{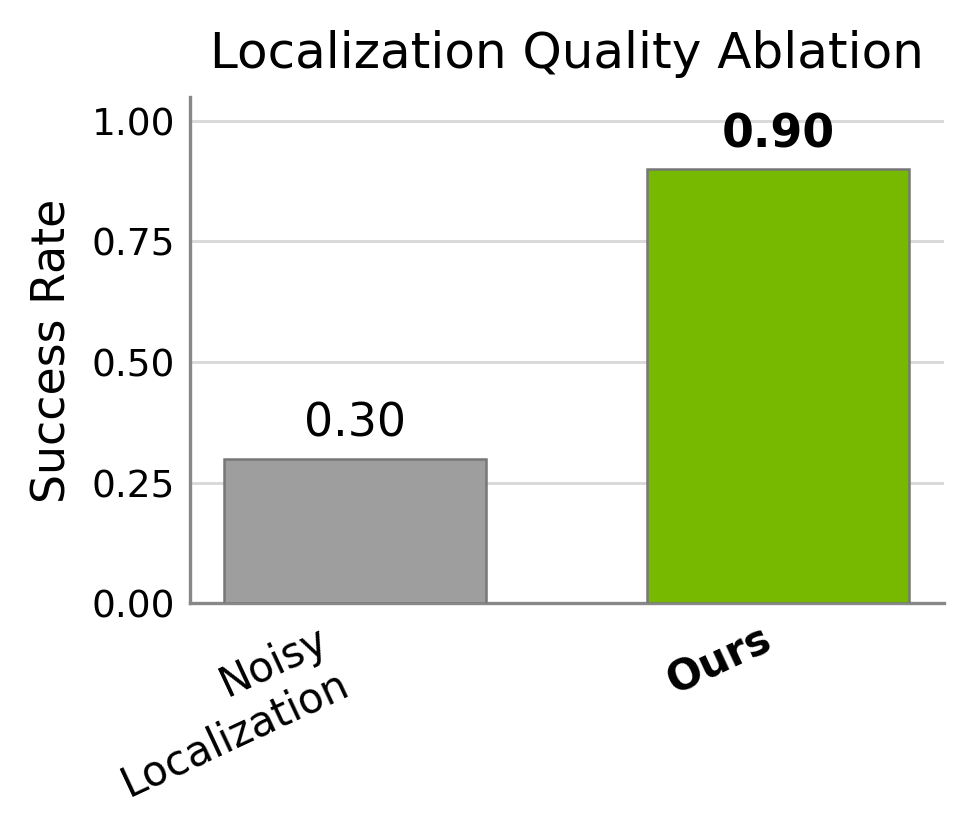}
  \caption{Policy success with synthetic localization noise compared with our original action labels.}
  \label{fig:localization_policy_success}
  \vspace{-0.8\baselineskip}
\end{wrapfigure}
\textbf{Effect of Localization Quality.} We aim to show that UMI policy success is directly correlated with localization quality. A UMI policy is inherently robust to absolute drift because its action space uses a relative trajectory representation: for an action sequence beginning at $t_0$, a sequence of $SE(3)$ transforms specifies the desired end-effector pose at time $t$ relative to its initial pose at $t_0$. This representation makes the system more robust to tracking errors during data collection. As a controlled experiment, we add synthetic drift noise---randomized offset biases at selected time steps in each localized trajectory---using the procedure detailed in Appendix~\ref{sec:appendix_synthetic_noise}. Adding this synthetic localization error increases the training-set translational ATE from $\sim$\SI{10}{\milli\meter} to more than \SI{30}{\milli\meter}. As shown in \figref{fig:localization_policy_success}, the resulting policy's success rate drops from \SI{90}{\percent} to \SI{30}{\percent}, with frequent grasp slippage and inaccurate object placement that causes collisions with the scene. This result quantitatively connects the quality of UMI action labels to downstream policy performance.

\subsection{Real-to-Sim Deployment}
\label{subsec:results_real2sim}

\begin{figure}[!t]
  \centering
  \begin{tabular}{@{}c@{\hspace{1pt}}c@{\hspace{1pt}}c@{\hspace{1pt}}c@{\hspace{1pt}}c@{}}
    \rotatebox{90}{\footnotesize Real} &
    \includegraphics[width=0.20\linewidth]{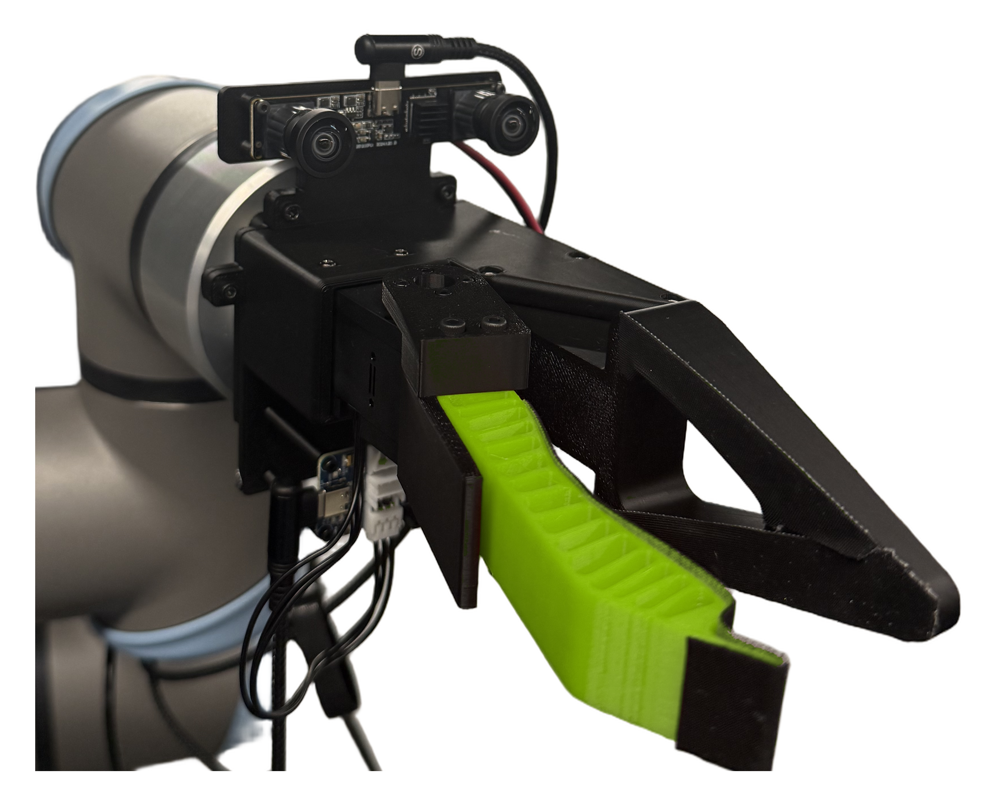} &
    \includegraphics[width=0.20\linewidth]{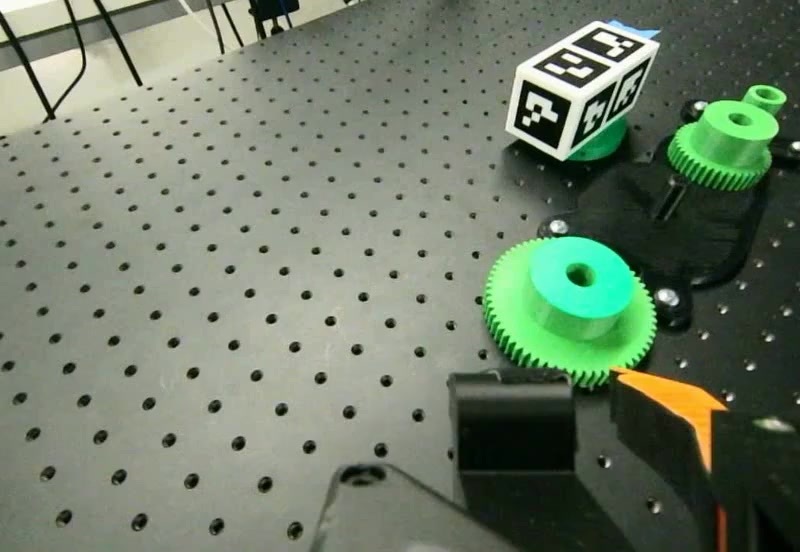} &
    \includegraphics[width=0.20\linewidth]{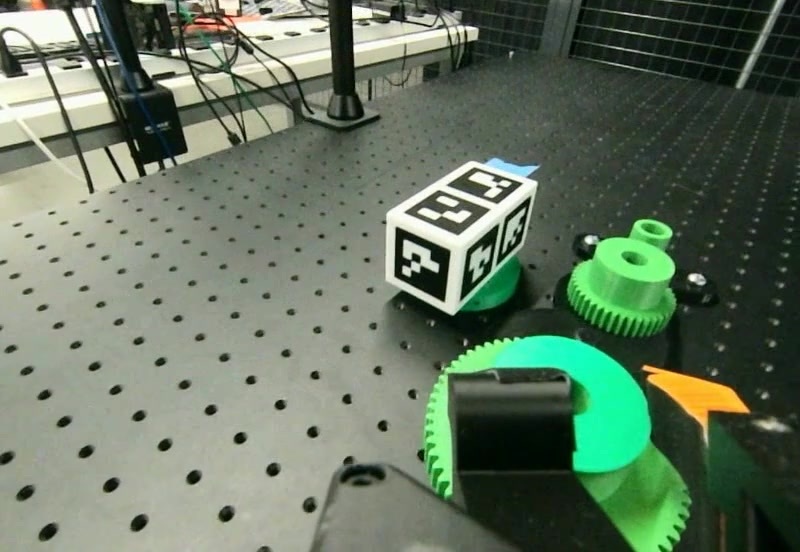} &
    \includegraphics[width=0.20\linewidth]{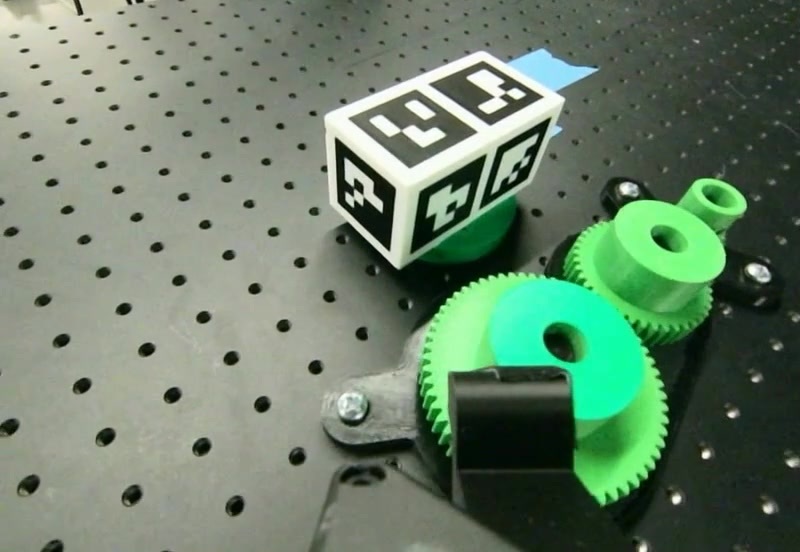} \\[-4pt]
    \rotatebox{90}{\footnotesize Sim} &
    \raisebox{1mm}{\includegraphics[width=0.20\linewidth]{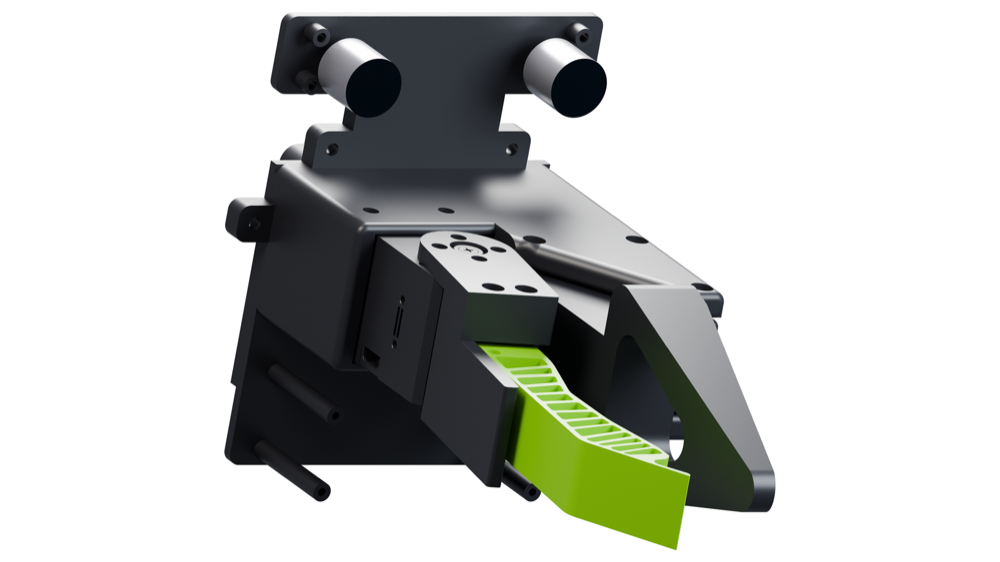}} &
    \includegraphics[width=0.20\linewidth]{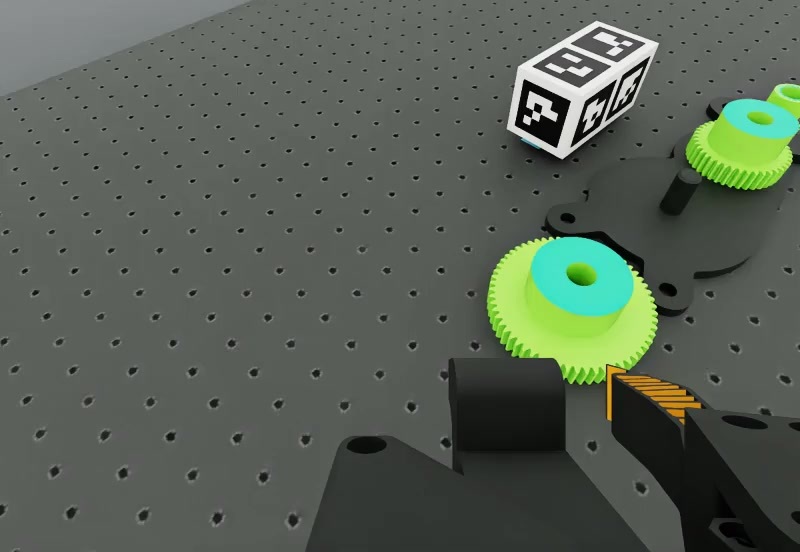} &
    \includegraphics[width=0.20\linewidth]{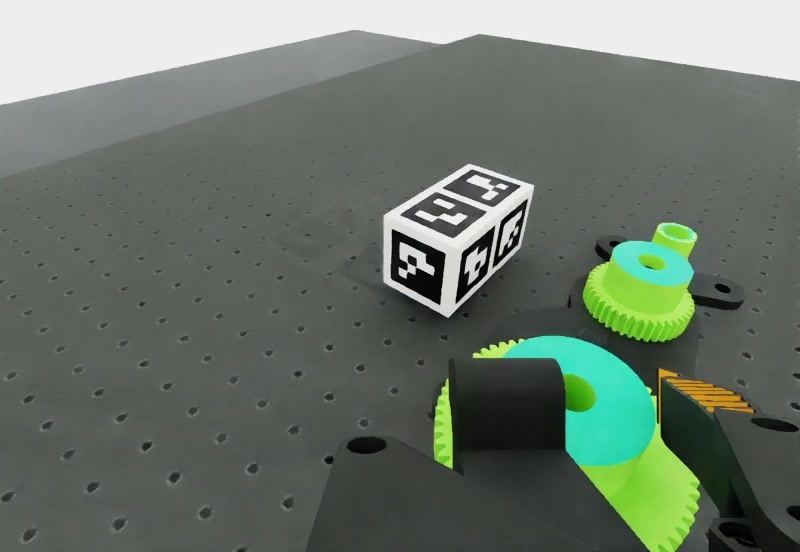} &
    \includegraphics[width=0.20\linewidth]{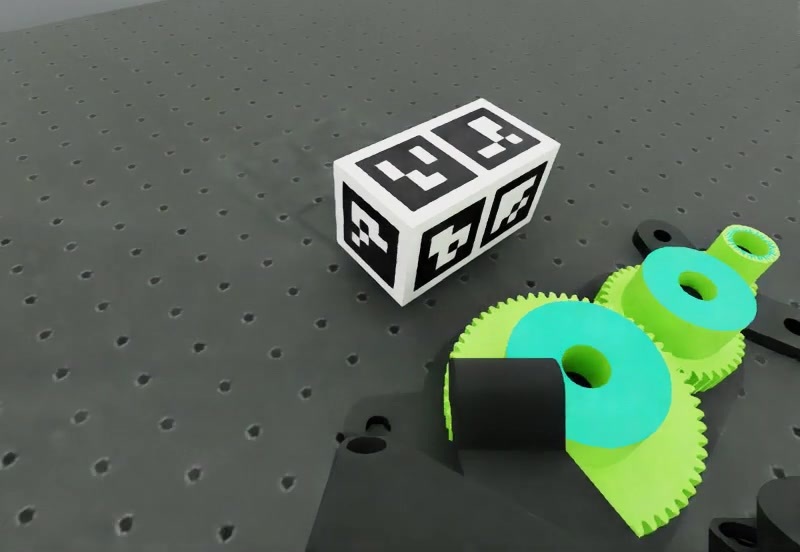} \\
  \end{tabular}
  \caption{\textbf{Agentically generated real-to-sim digital twin.} Corresponding gear insertion demonstration \figtop and successful Isaac Lab playback \figbottom; the agent recovers scene layout and contact geometry without ground-truth object poses.}
  \label{fig:real2sim}
\end{figure}

\ourmethod can also turn UMI demonstrations into simulation digital twins. Such environments provide repeatable settings for policy evaluation and can serve as a starting point for subsequent sim-to-real policy transfer. We generate these environments agentically with GPT-6 Astra~\cite{openai2026gpt6astra}. The agent receives only the UMI demonstration and the sim-ready assets for the UMI gripper and objects. Critically, it is not given the objects' ground-truth poses in the scene and must recover the scene graph from the demonstration.

We evaluate this agentic real-to-sim pipeline on all 10 demonstrations of the gear insertion task. The pipeline creates successful digital twins for 7 of 10 demonstrations, with success requiring the recorded task to play back successfully in simulation. Gear insertion is a contact-rich task whose simulation requires a signed-distance-field (SDF) collision representation; the agent recovers this representation automatically. \figref{fig:real2sim} shows one of the successful digital twins and its playback in the GPU-accelerated Isaac Lab simulator~\cite{mittal2025isaaclab}.

\FloatBarrier

\section{Conclusion}
\label{sec:conclusion}

We presented \ourmethod, an open hardware and software framework that combines a modular UMI-style gripper with stereo localization. The framework supports real-robot policy learning and agentic real-to-sim reconstruction, while UMI-Loc-Bench-v1 enables quantitative localization evaluation. Future work could validate the framework on bimanual tasks, which would require jointly localizing two independently moving stereo rigs. Another direction is to train higher capacity foundation model policies jointly on real and simulated data collected with the gripper.



\appendix
\begingroup
\setlength{\abovedisplayskip}{1pt plus 1pt minus 1pt}
\setlength{\belowdisplayskip}{5pt plus 1pt minus 1pt}
\setlength{\abovedisplayshortskip}{0pt plus 1pt}
\setlength{\belowdisplayshortskip}{1pt plus 1pt minus 1pt}
\section{Evaluation Metrics and Experimental Details}
\label{sec:appendix}
\vspace{-0.45\baselineskip}

In this appendix, we provide further details on the following: (A)~evaluation metrics, (B)~benchmark task specifications, (C)~the localization-error ablation, and (D)~the bill of materials.

\subsection{Evaluation Metrics}
\label{sec:appendix_metrics}
\vspace{-0.3\baselineskip}

\looseness=-1
We adopt the trajectory-evaluation metrics of the TUM RGB-D benchmark~\cite{sturm2012benchmark}. Let $P_1, \dots, P_n \in SE(3)$ denote the estimated camera poses and $Q_1, \dots, Q_n \in SE(3)$ the corresponding ground-truth poses, associated by timestamp.

\subsubsection{Absolute Trajectory Error}
\label{subsec:appendix_ate}
\vspace{-0.15\baselineskip}

\looseness=-1
Because the estimated and ground-truth trajectories are expressed in arbitrary coordinate frames, they must first be aligned. We compute the rigid-body transformation $S$ in closed form by Umeyama alignment~\cite{umeyama1991}, a least-squares SVD solution over all matched frame centers that maps the estimated trajectory onto the ground-truth frame, with the scale factor pinned to unity. The absolute trajectory error at time step $i$ is $F_i := Q_i^{-1} S P_i$ and is summarized as the root mean squared error over the translational components,
\begin{equation}
  \mathrm{RMSE}(F_{1:n}) := \left( \frac{1}{n} \sum_{i=1}^{n} \left\| \mathrm{trans}(F_i) \right\|^2 \right)^{1/2}.
  \label{eq:ate_rmse}
\end{equation}

\subsubsection{Rotational Absolute Trajectory Error}
\label{subsec:appendix_ate_rotation}
\vspace{-0.15\baselineskip}

\looseness=-1
Rotational ATE reuses the aligned residuals $F_i$ but scores their rotational component. For a rotation matrix $R$, the geodesic angle is $\angle(R) := \arccos\!\left((\mathrm{tr}(R)-1)/2\right)$, and rotational ATE is the root mean squared value of $\angle(\mathrm{rot}(F_i))$ over all frames, reported in degrees. Before scoring, we subtract a constant frame-convention offset, computed as the chordal mean of the residual rotations, so that a fixed misalignment between the sensor and ground-truth conventions is not charged as drift.

\vspace{-0.4\baselineskip}
\subsection{Benchmark Tasks and Protocol}
\label{sec:appendix_experiments}
\vspace{-0.3\baselineskip}

\looseness=-1
\benchmark contains 90 handheld demonstrations, with 10 demonstrations for each of nine tasks: block stacking, gear insertion, wrench rotation, shelf retrieval, bin retrieval, kitting, USB insertion, pen placement, and candy picking. These tasks span short-horizon pick-and-place, constrained retrieval, contact-rich insertion and non-prehensile manipulation across multiple scenes. A static cube carrying ten \SI{40}{\milli\meter} ArUco markers from the \texttt{DICT\_4X4\_100} dictionary defines the reference frame for each task. We estimate the camera pose from the visible marker corners and mask the cube in both camera streams before running any localization method, preventing the evaluated methods from using the ground-truth markers as tracking features.


\vspace{-0.3\baselineskip}
\subsection{Synthetic Localization Noise}
\label{sec:appendix_synthetic_noise}
\vspace{-0.2\baselineskip}

\looseness=-1
To isolate the effect of localization quality on policy learning, we synthetically perturb the translational components of the action-label trajectories to simulate tracking drift. For each episode, we seed the random-number generator and advance through the trajectory in intervals sampled uniformly from 10 to 30 frames. At the end of each interval, with probability 0.08, we sample a three-dimensional offset uniformly from a ball of radius \SI{10}{\centi\meter}. Each sampled offset is added to the current pose and all subsequent poses, so multiple injections accumulate as a piecewise-constant drift bias. 

\vspace{-0.3\baselineskip}
\subsection{Bill of Materials}
\label{sec:appendix_bom}
\vspace{-0.2\baselineskip}

\looseness=-1
Table~\ref{tab:bom} itemizes the components needed to build one \ourmethod gripper, using US retail prices as of October 2026.

\begin{table}[h]
  \centering
  \caption{\textbf{\ourmethod bill of materials.} Components for one deployment gripper along with its data-collection parts. Prices in USD; each component links to its source.}
  \label{tab:bom}
  \footnotesize
  \begin{tabularx}{\columnwidth}{X c r}
    \toprule
    Component & Qty & Price \\
    \midrule
    \multicolumn{3}{l}{\textit{Core Electronics}} \\
    \href{https://www.robotis.us/dynamixel-xc430-w150-t/}{Dynamixel XC430-W150-T actuator} & 1 & 137.89 \\
    \href{https://www.amazon.com/dp/B0H4GKGTWN}{MMlove USB global-shutter stereo camera} & 1 & 99.99 \\
    \href{https://www.robotis.us/openrb-150/}{ROBOTIS OpenRB-150 controller} & 1 & 28.64 \\
    \href{https://www.amazon.com/dp/B0DSZ8GQCM}{KY-023 joystick module (10 pack)} & 1 & 17.39 \\
    \midrule
    \multicolumn{3}{l}{\textit{Wiring}} \\
    \href{https://www.amazon.com/dp/B0FFN2B6CY}{Copper wire} & 1 & 22.99 \\
    \href{https://www.amazon.com/dp/B0G7GLKT73}{USB-A to USB-C cable, 10\,ft (2-pack)} & 1 & 20.99 \\
    \href{https://www.amazon.com/dp/B0BTT31CBC}{Jumper wire kit} & 1 & 11.99 \\
    \href{https://www.amazon.com/dp/B079RCNNCK}{5.5\,mm DC female power jacks} & 1 & 8.99 \\
    \href{https://www.amazon.com/dp/B07TPS44SL}{USB-C to dual USB-A adapter (connecting to laptop)} & 1 & 8.99 \\
    \href{https://www.amazon.com/dp/B0DYP7CFZZ}{Wire stripper} & 1 & 7.99 \\
    \midrule
    \multicolumn{3}{l}{\textit{Power \& Assembly}} \\
    \href{https://www.amazon.com/dp/B0BRWCFD6Q}{M3 socket-head screw kit} & 1 & 19.99 \\
    \href{https://www.amazon.com/dp/B01GEA8PQA}{12\,V 5\,A power supply} & 1 & 12.99 \\
    \midrule
    \textbf{Total} & & \textbf{398.83} \\
    \bottomrule
  \end{tabularx}
\end{table}


\endgroup

\ifsubmission
\else
\section*{ACKNOWLEDGMENT}
We want to thank Ken Goldberg, Yichao Pan, and Stephen Tyree for discussions and Michael Silverman for feedback on the hardware design.
\fi

\bibliographystyle{IEEEtran}
\bibliography{\iffullrefs references_full\else references\fi}

\end{document}